\documentclass[conference]{IEEEtran}
\usepackage[dvipsnames,svgnames,x11names]{xcolor} 
\usepackage{times}

\usepackage[title]{appendix}
\usepackage{multicol}
\usepackage{inconsolata} 
\usepackage{subcaption}

\usepackage{amsthm}
\theoremstyle{plain}

\theoremstyle{definition}

\theoremstyle{remark}

 \makeatletter
\def\@fnsymbol#1{\ensuremath{\ifcase#1\or \dagger\or \ddagger\or
  \mathsection\or \mathparagraph\or \|\or **\or \dagger\dagger
  \or \ddagger\ddagger \else\@ctrerr\fi}}
\makeatother

\usepackage{dsfont}

\usepackage{multirow}
\usepackage{amsthm,amssymb,lipsum}
\usepackage{times}
\usepackage{mathrsfs}
\usepackage{enumerate}
\usepackage{colortbl}
\usepackage{wrapfig}
\usepackage{bbding}
\usepackage{pifont}
\usepackage{xspace}
\usepackage{amsmath}
\usepackage{wasysym}
\usepackage{textcomp}
\usepackage{microtype}      
\usepackage[bottom]{footmisc}

\usepackage{makecell}

\usepackage{color}
\usepackage{tocloft}
\usepackage{caption}
\usepackage{float}
\usepackage{afterpage}

\usepackage{stackengine}

\usepackage{changes}

\usepackage{bbm}

\usepackage{cuted}
\usepackage{duckuments}

\usepackage{tabularx}

\usepackage[percent]{overpic}
\usepackage{adjustbox}
\usepackage{utfsym}

\def\eg{{\it e.g.}\xspace}

\def\ie{{\it i.e.}\xspace}
\def\etc{{\it etc}\xspace}

\definecolor{drp-blue}{HTML}{1f77b4}
\definecolor{pretty-blue}{RGB}{0, 113, 188}
\definecolor{kaiming-green}{RGB}{57,181,74} 
\definecolor{mypurple}{RGB}{55,0,168} 
\definecolor{icmlblue}{rgb}{0,0.08,0.45} 
\definecolor{linecolor1}{HTML}{F1F7FB}
\definecolor{linecolor2}{HTML}{E3EFF7}
\definecolor{linecolor3}{HTML}{D5E4F0}

\definecolor{reconcolor}{HTML}{412F8A}
\definecolor{runpei-orange}{HTML}{F35F27}
\definecolor{runpei_blue}{HTML}{14294B}
\definecolor{datacolor}{HTML}{0009BF}
\definecolor{vitcolor}{HTML}{fc8e62}

\newcommand{\sone}{Stage I\xspace}
\newcommand{\stwo}{Stage II\xspace}

\newcommand{\bh}{\bs{h}}
\newcommand{\bF}{\bs{F}}

\newcommand{\torque}{{\bs{{\tau}}_{t}}}
\newcommand{\dofpos}{{\bs{{d}}_{t}}}

\newcommand{\dofvel}{{\bs{\dot{d}}_{t}}}

\newcommand{\dofacc}{{\bs{\ddot{d}}_{t}}}

\newcommand{\bs}[1]{\boldsymbol{#1}}

\definecolor{runpei-orange}{HTML}{F35F27}
\definecolor{runpei-blue}{HTML}{14294B}

\usepackage[sort&compress,numbers]{natbib}

\definecolor{cvprblue}{rgb}{0.21,0.49,0.74}
\definecolor{myblue}{rgb}{.39,.58,.93}

\def\ours{{\scshape HumanUP}\xspace}

\usepackage{graphicx}
\usepackage{booktabs}

\usepackage{caption}
\usepackage{float}
\usepackage{colortbl}
\usepackage{lipsum}
\usepackage{changes}

\usepackage{appendix}
\usepackage{titletoc}

\usepackage[
    bookmarks=true,
    pagebackref,
    breaklinks,
    colorlinks,
    linkcolor=blue, 
    urlcolor=blue,
    citecolor=blue,
]{hyperref}
\usepackage[capitalize]{cleveref}
\Crefname{section}{Sec.}{Secs.}
\crefname{section}{Sec.}{Secs.}
\Crefname{table}{Tab.}{Tabs.}
\crefname{table}{Tab.}{Tabs.}
\Crefname{figure}{Fig.}{Figs.}
\crefname{figure}{Fig.}{Figs.}

\makeatletter
\def\blfootnote{\xdef\@thefnmark{}\@footnotetext}
\makeatother

\renewcommand{\thesection}{\Roman{section}}

\begin{document}

\title{Learning Getting-Up Policies for \\ Real-World Humanoid Robots}

\author{
\authorblockN{
Xialin He$^{\ast1}$\qquad
Runpei Dong$^{\ast1}$\qquad
Zixuan Chen$^2$\qquad
Saurabh Gupta$^1$
}
\vspace{0.05in}
\authorblockA{
$^1$University of Illinois Urbana-Champaign\qquad
$^2$Simon Fraser University
}
}  


\twocolumn[{%
\renewcommand\twocolumn[1][]{#1}%
\maketitle
\begin{center}
    \centering
    \captionsetup{type=figure}
    \includegraphics[width=\linewidth]{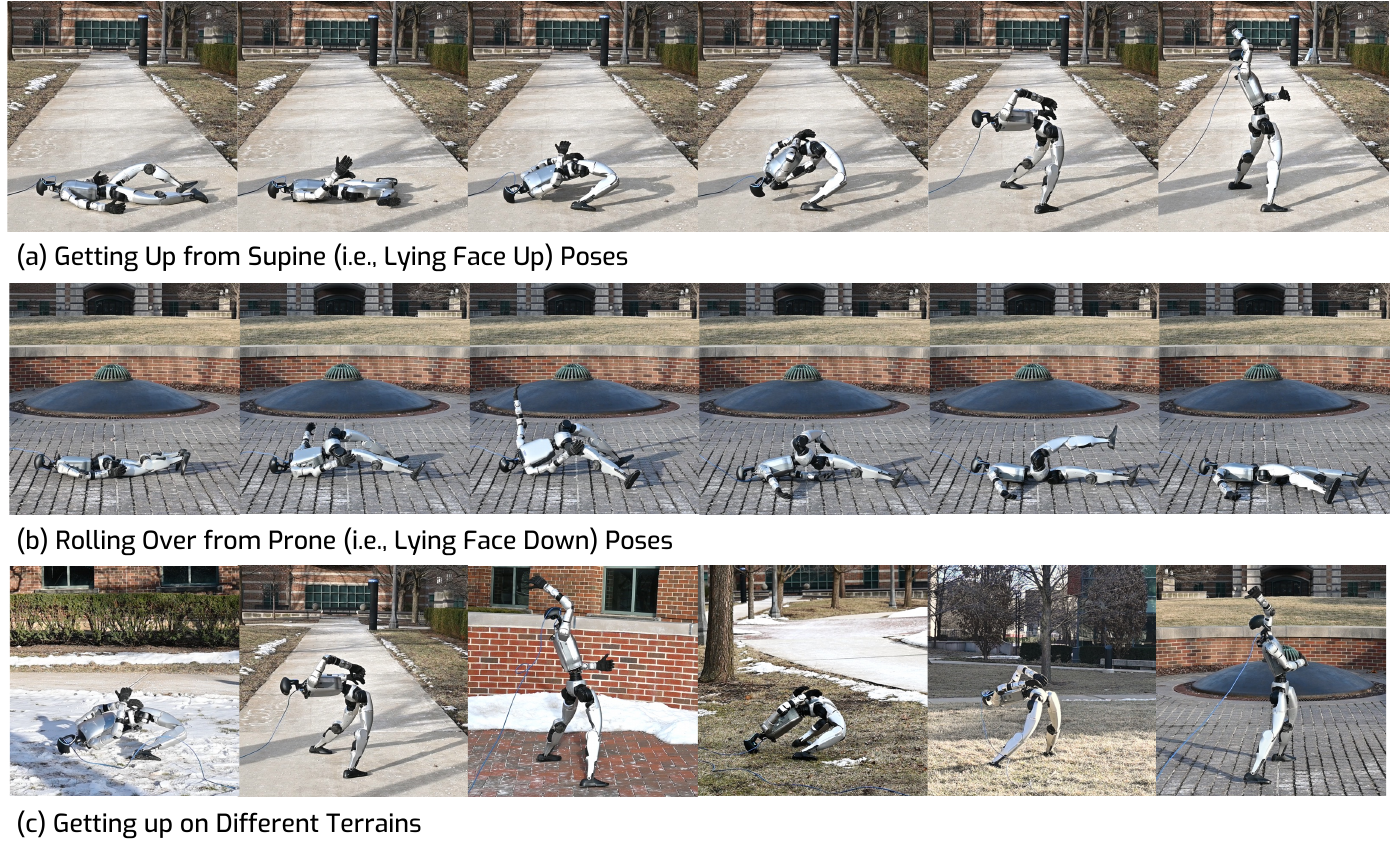}
    \caption{\ours provides a simple and general two-stage training method for humanoid getting-up tasks, which can be directly deployed on Unitree G1 humanoid robots~\cite{UnitreeG124}. Our policies showcase robust and smooth behavior that can get up from diverse lying postures (both supine and prone) on varied terrains such as grass slopes and stone tiles. 
    }\label{fig:teaser}
\end{center}
}]

{\blfootnote{{$^{\ast}$ Equal contributions.}}}

\begin{abstract}
Automatic fall recovery is a crucial prerequisite before humanoid robots can be reliably deployed. Hand-designing controllers for getting up is difficult because of the varied configurations a humanoid can end up in after a fall and the challenging terrains humanoid robots are expected to operate on. This paper develops a learning framework to produce controllers that enable humanoid robots to get up from varying configurations on varying terrains. Unlike previous successful applications of learning to humanoid locomotion, the getting-up task involves complex contact patterns (which necessitates accurately modeling of the collision geometry) and sparser rewards. We address these challenges through a two-phase approach that induces a curriculum. The first stage focuses on discovering a good getting-up trajectory under minimal constraints on smoothness or speed / torque limits. The second stage then refines the discovered motions into deployable (\ie smooth and slow) motions that are robust to variations in initial configuration and terrains. 
We find these innovations enable a real-world G1 humanoid robot to get up from two main situations that we considered: 
a) lying face up and b) lying face down, both tested on flat, deformable, slippery surfaces and slopes (\eg, sloppy grass and snowfield).
This is one of the first successful demonstrations of learned getting-up policies for human-sized humanoid robots in the real world.
\\Project page: \href{humanoid-getup.github.io}{\url{https://humanoid-getup.github.io/}}
\end{abstract}


\section{Introduction}
This paper develops learned controllers that enable a humanoid robot to get up
from varied fall configurations on varied terrains.  Humanoid robots are
susceptible to falls, and their reliance on humans for fall recovery hinders their
deployment. Furthermore, as humanoid robots are expected to work in
environments involving complex terrains and tight workspaces (\ie challenging
scenarios that are too difficult for wheeled robots), a humanoid robot may end up
in an unpredictable configuration upon a fall, or may be on an unknown terrain.
26 of the 46 trials at
the DARPA Robotics Challenge (DRC) had a fall, and 25 of these falls required
human intervention for recovery~\cite{krotkov2018darpa}. The DRC identified
fall prevention and recovery as a major topic needing more research.  This
paper pursues it and proposes a learning-based framework for
learning fall recovery policies for humanoid robots under varying
conditions.

The need for recovering from varied initial conditions makes it hard to design a fall recovery controller by hand and motivates the need for learning via trial
and error in simulation. Such learning has produced exciting results
in recent years for locomotion problems involving quadrupeds and
humanoids, \eg~\cite{AgileDynamicMotorSkills19,RealWorldHumanoidLocomotionScienceRobotics24}.
Motivated by these exciting results, we started with simply applying the Sim-to-Real (Sim2Real) paradigm
for the getting-up problem. However, we quickly realized that the getting-up problem is
different from typical locomotion problems in the following three significant
ways that made a naive adaptation of previous work inadequate: 

\begin{itemize}
    \item[\textbf{a)}] \textbf{Non-periodic behavior.} 
    In locomotion, contacts with the environment happen in structured ways:
cyclic left-right stepping pattern. The getting-up problem doesn't have such a periodic behavior. The contact sequence necessary for getting up
itself needs to be figured out. This makes optimization harder and may render
phase coupling of left and right feet commonly used in locomotion
ineffective. 
\item[\textbf{b)}] \textbf{Richness in contact.} Different from locomotion, 
contacts necessary for getting up are not
limited to just the feet. Many other parts of the robot are likely already in
touch with the terrain. But more importantly, the robot may find it useful to
employ its body, outside of the feet, to exert forces upon the environment, in
order to get up. Freezing / decoupling the upper body, only coarsely modeling
the upper body for collisions, and using a larger simulation step size: the
typical design choices made in locomotion, are no longer applicable for the
getting up task.
\item[\textbf{c)}] 
\textbf{Reward sparsity.} Designing rewards for getting up is harder than other locomotion tasks. 
Velocity tracking offers a dense reward and feedback on
whether the robot is meaningfully walking forward is available within a few tens
of simulation steps. In contrast, many parts of the body make negative
progress, \eg, the torso first needs to tilt down for seconds before
tilting up to finally get up. 
\end{itemize}

We present \ours, a two-stage reinforcement learning (RL) training framework that circumvents these issues.
\sone targets \textit{solving the task} in easier settings (sparse task rewards with weak regularization), while \stwo makes the learned motion \textit{deployable} (\ie, control should be smooth; velocities and executed torques should be small; \etc).
Discovering the getting-up motion is hard because of sparse and underspecified rewards. \sone tackles this hard problem without being limited by smoothness in motion or speed / torque limits.
Tracking a trajectory is easier as it offers dense rewards. \stwo tackles this easier problem but does it under strict Sim2Real control regularization and randomization of terrains and initial poses.
Thus, going from \sone to \stwo corresponds to a
learning curriculum that progresses from
simplified $\rightarrow$ full collision mesh, canonical $\rightarrow$ random initial lying posture,  and weak $\rightarrow$ strong control regularization, and domain randomization.
This amounts to a \textit{hard-to-easy} curriculum on task difficulty (Stage
I: getting-up task; Stage II: motion tracking), and an \textit{easy-to-hard} curriculum on regularization and variability (Stage I: weaker, Stage II: stronger).

We conduct experiments in simulation and the real world with the G1 platform
from Unitree. In the real world, we find our framework enables the G1 robot to
get up from two different poses (supine, \ie lying face up, and prone, \ie lying face down) across six different terrains.
This expands the capability of the G1 robot: the manufacturer-provided hand-crafted getting-up controller only successfully gets up from supine poses 
on a flat surface without bumps. In simulated experiments, our framework can
successfully learn getting-up policies that work on varied terrains and varied
starting poses.

\section{Related Work}
We review related works on humanoid control, learning for humanoid control, and work specifically targeted toward fall recovery for legged robots.

\subsection{Humanoid Control}
Controlling a high degree of freedom humanoid robots has fascinated researchers for the last several decades. 
Model-based techniques, such as those based on the Zero Moment Point (ZMP) principle~\cite{ZeroMoment04,HondaHumanoid98,ASIMO02,WalkMan17}, optimization~\cite{OptimizationBasedLocomotion16,bookwalkingrunning2023,BipedalRunning23}, and Model Predictive Control (MPC)~\cite{MITHumanoid21,galliker2022planar,DynamicLocomotionMITConvexMPC18,FullOrderSamplingBasedMPC24}, have demonstrated remarkable success in fundamental locomotion tasks like walking, running and jumping.
However, these approaches often struggle to generalize or adapt to novel environments.
In contrast, learning-based approaches have recently made significant strides, continuously expanding the generalization capabilities of humanoid locomotion controllers.

\subsubsection{Learning for humanoid control}
\begin{figure*}
  \centering
    \includegraphics[width=\linewidth]{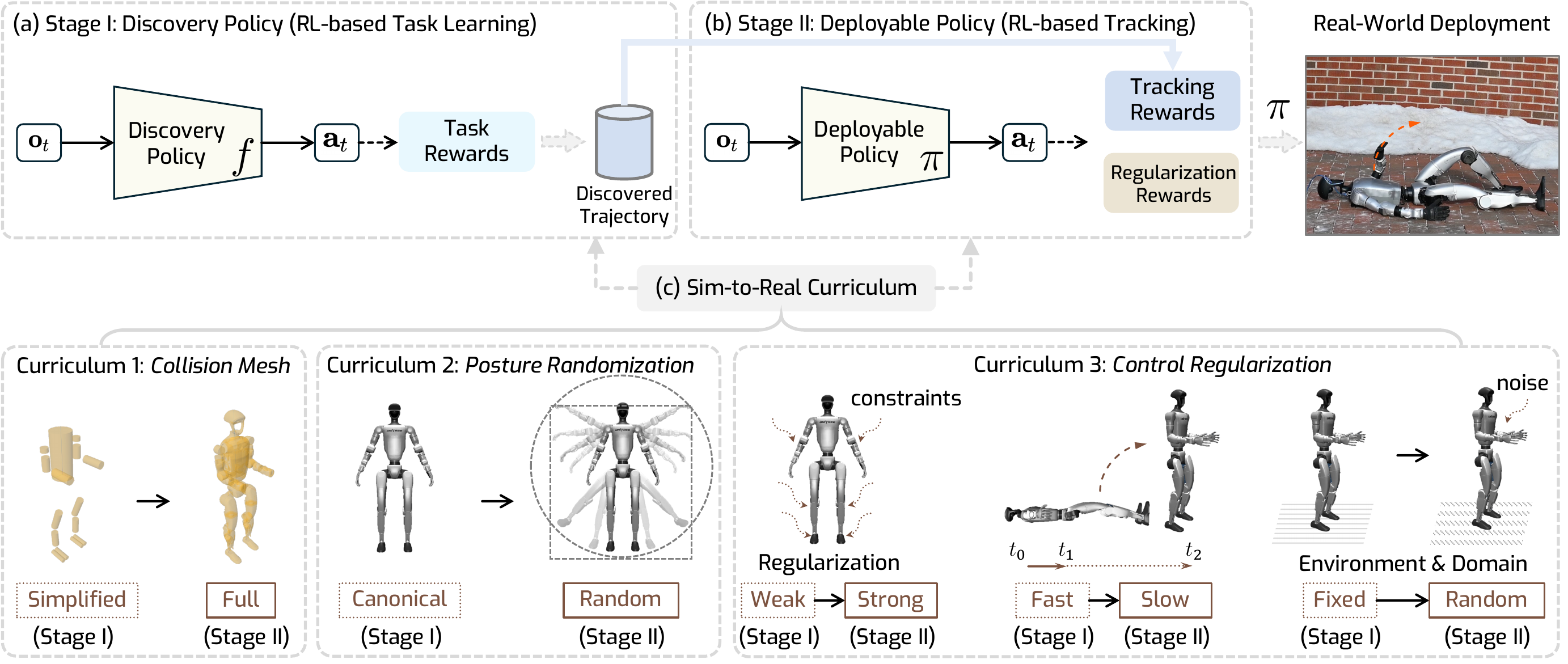}
  \caption{\textbf{\ours system overview}. 
  Our getting-up policy (\cref{sec:policy}) is trained in simulation using two-stage RL training, after which it is directly deployed in the real world. 
  (a) \sone (\cref{sec:stageI}) learns a discovery policy $f$ that figures out a getting-up trajectory with minimal deployment constraints.
  (b) \stwo (\cref{sec:stageII}) converts the trajectory discovered by \sone into a policy $\pi$ that is deployable, robust, and generalizable. This policy $\pi$ is trained by learning to track a slowed down version of the discovered trajectory under strong control regularization on varied terrains and from varied initial poses.
  (c) The two-stage training induces a curriculum (\cref{sec:curriculum}). \sone targets motion discovery in easier settings (simpler collision geometry, same starting poses, weak regularization, no variations in terrain), while \stwo solves the task of making the learned motion deployable and generalizable. 
  }\label{fig:overview}
  \vspace{-3pt}
\end{figure*}
Learning in simulation via reinforcement followed by a sim-to-real transfer has led to many successful locomotion results for quadrupeds~\cite{AgileDynamicMotorSkills19, RMA21} and humanoids~\cite{RealWorldHumanoidLocomotionScienceRobotics24, HumanoidLocomotionNextTokenPrediction24, HumanoidLocomotionChallengingTerrain24, advancinglocomotion2024, LCP24, KinodynamicFabrics23}. This has enabled locomotion on challenging in-the-wild terrain~\cite{HumanoidLocomotionChallengingTerrain24,DenoisingWorldModel24}, agile motions like jumping~\cite{BipedalJumpingControl23,WoCoCo24}, and even locomotion driven by visual inputs~\cite{HumanoidParkour24,long2024learning}. Researchers have also expanded the repertoire of humanoid motions to skillful movements like dancing and naturalistic walking gaits through use of human mocap or video data~\cite{Exbody2_24, Exbody24, UH1_24, Hover24}. Some works address locomotion and manipulation problems for humanoids simultaneously to enable loco-manipulation controllers in an end-to-end fashion facilitated by teleportation~\cite{OmniH2O24,HumanPlus24,MobileTelevision24}. 
Notably, these tasks mostly involve contact between the feet and the environment, thus requiring only limited contact reasoning. How to effectively develop controllers for more \textit{contact-rich} tasks like crawling, tumbling, and getting up that require numerous, dynamic, and unpredictable contacts between the whole body and the environment remains under-explored.

\subsection{Legged robots fall recovery}
Humanoid robots are vulnerable to falls due to under-actuated control dynamics, high-dimensional states, and unstructured environments~\cite{WABOT1_73,HondaHumanoid98,MABEL09,krotkov2018darpa,Humanoid35DoF96,gu2025humanoid}, making the ability to recover from falling of great significance. 
Over the years, this problem has been tackled in the following ways.

\subsubsection{Getting up via motion planning}
Early work from \citet{Learning2StandUp98} solved the getting-up problem for a two-joint, three-link walking robot in 2D, and several discrete states are used as subgoals to transit via hierarchical RL. 
This line of work can be viewed as an application of motion planning by \textit{configuration graph transition} learning~\cite{StateTransitionGraph96}, where stored robot states between lying and standing are used as graph nodes to transit~\cite{UKEMI02,FirstHumanoidGetUp03,GettingUpMotionPlanning07,HumanoidBalancing16}.
More recently, some progress has been made to enable toy-sized humanoid robots to get up~\cite{HumanoidStandingUpLfDMultimodalReward13,StandUpSymmetry16,FastAndReliable21,FRASA24}.
For example, \citet{HumanoidStandingUpLfDMultimodalReward13} explores getting up from a canonical sitting posture with motion planning by imitating human demonstration with ZMP criterion.
To address the high-dimensionality of humanoid configurations, \citet{StandUpSymmetry16} leverage bilateral symmetry to reduce the control DoFs by half and a clustering technique is used for further reducing the complexity of configuration space, thereby improving getting-up learning efficiency.
However, such state machine learning using predefined configuration graphs may not be sufficient for generalizing to unpredictable initial and intermediate states, which happens when the robot operates on challenging terrains.

\subsubsection{Hand-designed getting-up trajectories} 
Another solution, often adopted by commercial products, is to replay a manually designed motion trajectory. For example, Unitree~\cite{UnitreeG124} has a getting-up controller built into G1's default controllers. Booster Robotics~\cite{Booster} designed a specific recovery controller for their robots that can help the robot recover from fallen states. 
Concurrent work from Zhuang and Zhao~\cite{EmbraceCollisions25} enables a G1 robot to get up by tracking the getting-up motion of a real human.
The main drawback of such pre-defined trajectory getting-up controllers is that they may only handle a limited number of fallen states and lack generalization.

\subsubsection{Learned getting-up policies for real robots} 
RL followed by sim-to-real has also been successfully applied for quadruped~\cite{AgileDynamicMotorSkills19,dribblebot2023,guardiansasyoufall2024,ma2023learning} fall recovery. 
For example, \citet{AgileDynamicMotorSkills19} explore sim2real RL to achieve real-world quadruped fall recovery from complex configurations.
\citet{dribblebot2023} train a recovery policy that enables the quadruped to dribble in snowy and rough terrains continuously.
\citet{guardiansasyoufall2024} develop a quadruped recovery policy in highly dynamic scenarios.

\subsubsection{Learned getting-up policies for character animation} A parallel research effort in character animation, also explores the design of RL-based motion imitation algorithms: DeepMimic~\cite{DeepMimic18}, AMP~\cite{AMP2021}, PHC~\cite{PHC23}, among others~\cite{PhysicsBasedMocapImitationDRL18,PULSE24,MaskedMimic24,HierachicalWorldModel25,CooHOI24,HOIHumanLevelInstruction24}. These have also demonstrated successful getting-up controllers in simulation. 
By tracking user-specified getting-up curves, \citet{frezzato2022synthesizing} enable humanoid characters to get up by synthesizing physically plausible motion.
Without recourse to mocap data, such naturalistic getting-up controllers for simulated humanoid characters can also be developed with careful curriculum designs~\cite{Learning2GetUp22}.
Some works explore sampling-based methods for addressing contact-rich character locomotion, including getting up~\cite{SAMCONSamplingBasedContactRichMotion10,OnlineMotionSynthesis14,SampleEfficientCE21}, while some works have demonstrated success in humanoid getting up with online model-predictive control~\cite{OnlineTrajectoryOptimization12}.
It is worth noticing, however, that these works use humanoid characters with larger DoFs compared to humanoid robots (\eg, 69 DoFs in SMPL~\cite{SMPL15}) and use simplified dynamics.
As a result, learned policies operate body parts at high velocities and in infeasible ways, leading to behavior that cannot be transferred into the real world directly.
Hence, developing generalizable recovery controllers for humanoid robots remains an open problem. 

\section{\ours: Sim-to-Real Humanoid Getting Up}
Our goal is to learn a getting-up policy $\pi$ that enables a humanoid to get up from arbitrary initial postures. We consider getting up from two families of lying postures: a) supine poses (\ie lying face up) and b) prone poses (\ie lying face down).
Getting up from these two groups of postures may require different behaviors, which makes it challenging to learn a single policy that handles both scenarios.
To tackle this issue, we decompose the getting-up task from a prone pose to first rolling over and then standing up from the resulting supine posture. Therefore, we aim to learn policies for rolling over from a prone pose and getting up from a supine pose separately.

To solve these two tasks, we propose \ours, a general learning framework for training getting-up and rolling over policies, which is illustrated in \cref{fig:overview}.
In \sone, a discovery policy $f$ is trained to figure out standing-up or rolling-over motions.
$f$ is trained without deployment constraints, and only the task and symmetry rewards are used.
In stage II, a deployable policy $\pi$ imitates the rolling-over / getting-up behaviors obtained from stage I under strong control regularization.
This deployable policy $\pi$ is transferred from simulation to the real world as the final policy.
We detail the policy model and two-stage training in \cref{sec:two-stage-policy}, and then discuss the induced curriculum in \cref{sec:curriculum}.

\subsection{Policy Architecture}\label{sec:policy}
\ours trains two policy models $f$ and $\pi$ with RL.
Both policy models take observation $\mathbf{o}_t = [\mathbf{z}_t, \mathbf{s}_t, \mathbf{s}_{t-10:t-1}] \in \mathbb{R}^{868}$ as input and output action $\mathbf{a}_t \in \mathbb{R}^{23}$, where $\mathbf{s}_t \in \mathbb{R}^{74}$ is the proprioceptive information, $\mathbf{s}_{t-10:t-1}$ is the $10$ steps history states.
$\mathbf{z}_t\in\mathbb{R}^{54}$ are the encoded environment extrinsic latents that are predicted from observation history and learned using regularized online adaptation~\cite{DeepWholeBodyControl22}.
The proprioceptive information $\mathbf{s}_t$ consists of the robot's \texttt{roll} and \texttt{pitch}, angular velocity, DoF velocities, and DoF positions.
Such proprioceptive information can be accurately obtained in the real world, and we find that this is sufficient for the robot to infer the overall posture.
We do not use any linear velocity and \texttt{yaw} information as it is difficult to reliably estimate them in the real world~\cite{Human2HumanTeleoperation24,OmniH2O24}. 

The policy models are implemented as MLPs and trained via PPO~\cite{PPO17}.
The optimization maximizes the expected $\gamma$-discounted policy return within $T$ episode length: 
\(\mathbb{E}\left[\sum_{t=1}^T \gamma^{t-1}r_t\right]\), where $r_t$ is the reward at timestamp $t$.

\subsection{Two-Stage Policy Learning}\label{sec:two-stage-policy}

\subsubsection{\sone: Discovery Policy}\label{sec:stageI}
This stage discovers getting-up / rolling-over behavior efficiently without deployment constraints. 
We use the following task rewards with very weak regularization to train this discovery policy $f$. 
Timestep $t$ and reward weight terms are omitted for simplicity. The precise expressions for each reward term and their weights are provided in \cref{app:rewards_I}.

\vspace{5pt}
\noindent\textbf{Rewards for Getting Up:} $r_{\text{up}} = r_{\text{height}} + r_{\Delta \text{height}} + r_{\text{uprightness}} + r_{\text{stand\_on\_feet}} + r_{\Delta\text{feet\_contact\_forces}} + r_\text{symmetry}$, 
    where \\
    \textbullet~\(r_{\text{height}}\) encourages the robot's height to be close to a target height when standing; \\
    \textbullet~\(r_{\Delta\text{height}}\) encourages the robot to continuously increasing its height; \\
    \textbullet~\(r_{\text{uprightness}}\) encourages the robot to increase the z-component of the projected gravity,\footnote{Projected gravity on a robot part is the gravity vector transformed from the
    world frame to the part's local frame.} so that the robot stands upright; \\
    \textbullet~\(r_{\text{stand\_on\_feet}}\) encourages the robot to stand on both feet; \\
    \textbullet~\(r_{\Delta\text{feet\_contact\_forces}}\) encourages the robot to increase contact forces applied to the feet continuously;\\
    \textbullet~\(r_{\text{symmetry}}\) reduces the search space by {\it encouraging (but not requiring)} the robot to output bilaterally symmetric actions. Past work~\cite{SymmetricLeggedLocomotion24,StandUpSymmetry16} employed hard symmetry which improves RL sample efficiency at the cost of limiting robots' DoFs and generalization. Our {\it soft symmetry reward} partially leverages the benefit but mitigates the limitation.

\vspace{5pt}
\noindent\textbf{Rewards for Rolling Over:} ~\(r_{\text{roll}} = r_{\text{gravity}}\), which encourages the robot to change its body orientation so that its projected gravity is close to the projected gravity when lying face up.

\subsubsection{\stwo: Deployable Policy}\label{sec:stageII}
This stage trains policy $\pi$ that will be directly deployed in the real world. Policy $\pi$ is trained to imitate an $8\times$ slowed-down version of the state trajectories discovered in \sone, while also respecting strong regularization to ensure Sim2Real transferability. We use the typical regularization rewards and describe them in \cref{app:rewards_II}. Below, we describe the tracking reward.

\vspace{5pt}
\noindent\textbf{Tracking Rewards:}
$r_{\text{tracking}}$ encourages the robot to act close to the given motion trajectory derived from the discovered motion. \(r_{\text{tracking}} = r_{\text{tracking\_DoF}} + r_{\text{tracking\_body}}\), where \\
\textbullet~\(r_{\text{tracking\_DoF}}\) encourages the robot to move to the same DoF position as the reference motion, and \\
\textbullet~\(r_{\text{tracking\_body}}\) encourages the robot to move the body to the same position as the reference.
Specifically, $r_{\text{tracking\_body}}$ becomes two different rewards to encourage tracking upright posture ($r_{\text{head\_height}}$) and correct head orientation ($r_{\text{head\_gravity}}$) for getting-up and rolling-over tasks, respectively.

\subsection{\sone to \stwo Curriculum}\label{sec:curriculum}
The design of two-stage policy learning induces a
\textit{hard-to-easy} curriculum~\cite{CurriculumLearning09}. \sone targets
motion discovery in easier settings (weak regularization, no variations
in terrain, same starting poses, simpler collision geometry). Once motions have
been discovered, \stwo solves the task of making the learned motion deployable
and generalizable. As our experiments will show, splitting the work into two
phases is crucial for successful learning. Specifically, complexity increases
from \sone to \stwo in the following ways:

\subsubsection{Collision mesh}
As shown in \cref{fig:overview}, \sone uses a simplified collision mesh for faster motion discovery, while \stwo uses the full mesh for improved Sim2Real performance. 

\subsubsection{Posture randomization}

\sone learns to get up (and roll over) from a canonical pose, accelerating learning, while \stwo starts from arbitrary initial poses, enhancing generalization. To further speed up \sone, we mix in standing poses.
For \stwo, we generate a dataset \(\mathcal{P}\) of 20K supine poses \(\mathcal{P}_{\text{supine}}\) and 20K prone poses \(\mathcal{P}_{\text{prone}}\) by randomizing initial DoFs from canonical lying poses, dropping the humanoid from 0.5m, and simulating for 10s to resolve self-collisions. We use 10K poses from each set for training and the rest for evaluation.

\subsubsection{Control Regularization and Terrain Randomization}
For Sim2Real transfer, we use the following control regularization terms and environment randomization in \stwo: \\
\textbullet~\textbf{Weak $\rightarrow$ strong control regularization.} Weak control regularization in \sone enables discovery of getting-up / rolling-over motion, while strong control regularization (via smoothness rewards, DoF velocity penalties, \etc, see the full list in \cref{app:rewards_II}) in \stwo encourages more deployable action.
\\
\textbullet~\textbf{Fast $\rightarrow$ slow motion speed. }
    Without strong control regularization, \sone discovers a fast but unsafe getting-up motion ($<$1s), infeasible for real-world deployment. 
    To address this, we slow it to 8s via interpolation, providing stable tracking targets for \stwo, which better aligns with its control regularization.
\\  
\textbullet~\textbf{Fixed $\rightarrow$ random dynamics and domain parameters.}
    \stwo also employs domain and dynamics randomization via terrain randomization and noise injection. Such randomization has been shown to play a vital role in successful Sim2Real~\cite{DomainRandomizationSim2Real17}.

\begin{table*}[t!]
\caption{\textbf{Simulation results}. 
We compare \ours with several baselines on the \texttt{held-out} split of our curated posture set \(\mathcal{P}_{\text{supine}}\) and \(\mathcal{P}_{\text{prone}}\) using full URDF.
All methods are trained on the \texttt{training} split of our posture set $\mathcal{P}$, except for methods \ours w/o Stage II and w/o posture randomization.
\ours solves task \ding{184} by solving task \ding{183} and task \ding{182} consecutively.
We do not include the results of baseline 6 (\ours w/o Two-Stage Learning) as it cannot solve the task.
\(^\dagger\) \citet{Learning2GetUp22} is trained to directly solving task \ding{184} as it does not have a rolling over policy.
{\scshape Sim2Real} column indicates whether the method can transfer to the real world successfully. We tested all methods in the real world for which the {\scshape Smoothness} and {\scshape Safety} metrics are reasonable, and {\scshape Sim2Real} is false if deployment wasn't successful.
Metrics are introduced in \cref{sec:metrics}.
}\label{tab:simulation_results}
\centering
\resizebox{\linewidth}{!}{
\begin{tabular}{lccccccccc}
\toprule[0.95pt]
\multirow{2}{*}[-0.5ex]{} & \multirow{2}{*}[-0.5ex]{\scshape Sim2Real} & \multicolumn{2}{c}{{\scshape Task}} & \multicolumn{3}{c}{{\scshape Smoothness}} & \multicolumn{2}{c}{{\scshape Safety}}\\
\cmidrule(lr){3-4}
\cmidrule(lr){5-7}
\cmidrule(lr){8-9}
  & & Success$\uparrow$ & Task Metric$\uparrow$ & Action Jitter$\downarrow$ &  DoF Pos Jitter$\downarrow$ & Energy$\downarrow$ & $\mathcal{S}^{\text{Torque}}_{0.8, 0.5} \uparrow$ & $\mathcal{S}^{\text{DoF}}_{0.8, 0.5} \uparrow$\\ 
\midrule[0.6pt]

\multicolumn{9}{l}{\ding{182} \textit{\textbf{Getting Up from Supine Poses}}}\\
\midrule[0.6pt]
\citet{Learning2GetUp22} & \ding{55} & 92.62 $\pm$ 0.54  & \textbf{1.27 $\pm$ 0.00}  & 5.39 $\pm$ 0.01 & 0.48 $\pm$ 0.00 & 650.19 $\pm$1.26 & 0.72 $\pm$ 3.10e-4 & 0.73 $\pm$ 1.39e-4 \\

\ours w/o Stage II & \ding{55} & 24.82 $\pm$ 0.25 & 0.83 $\pm$ 0.00  & 13.70 $\pm$ 0.18 & 0.71 $\pm$ 0.00 & 1311.22 $\pm$ 8.57 & 0.57 $\pm$ 1.45e-3  & 0.67 $\pm$ 5.56e-4 \\

\ours w/o Full URDF & \ding{55} &  93.95 $\pm$ 0.24 & 1.22 $\pm$ 0.00 & 0.71 $\pm$ 0.00 & 0.11 $\pm$ 0.00 & 104.14 $\pm$ 0.57 & 0.92 $\pm$ 8.36e-5 & 0.77 $\pm$ 9.40e-5 \\

\ours w/o Posture Rand. & \ding{51} & 65.39 $\pm$ 0.50 & 1.09 $\pm$ 0.04 & 0.75 $\pm$ 0.05 & 0.15 $\pm$ 0.03 & 141.52 $\pm$ 0.61 & 0.91 $\pm$ 2.32e-4 & 0.74 $\pm$ 7.24e-5 \\

\ours w/ Hard Symmetry & \ding{51}  & 84.56 $\pm$ 0.11 & 1.23 $\pm$ 0.00 & 0.97 $\pm$ 0.01 & 0.22 $\pm$ 0.00 & 182.39 $\pm$ 0.22 & 0.89 $\pm$ 1.70e-5  & 0.78 $\pm$ 8.81e-5 \\
\ours & \ding{51} & \textbf{95.34 $\pm$ 0.12} & 1.24 $\pm$ 0.00 & \textbf{0.56 $\pm$ 0.01} &\textbf{0.10 $\pm$ 0.00} & \textbf{91.74 $\pm$ 0.33} & \textbf{0.93 $\pm$ 1.55e-5} & \textbf{0.78 $\pm$ 4.15e-5} \\
\midrule[0.6pt]

\multicolumn{9}{l}{\ding{183} \textit{\textbf{Rolling Over from Prone to Supine Poses}}}\\
\midrule[0.6pt]

\ours w/o Stage II & \ding{55} & 43.48 $\pm$ 0.41 & 0.91 $\pm$ 0.00  & 3.32 $\pm$ 0.31 & 0.40 $\pm$ 0.05  & 1684.66 $\pm$ 0.43 & 0.65 $\pm$ 6.28e-4  & 0.72 $\pm$ 7.18e-5 \\

\ours w/o Full URDF & \ding{55} & 87.73 $\pm$ 0.33 & 0.97 $\pm$ 0.00  & 0.33 $\pm$ 0.00 & 0.07 $\pm$ 0.00 & 59.01 $\pm$ 0.05 & 0.93 $\pm$ 7.91e-5  & 0.75 $\pm$ 9.98e-5 \\

\ours w/o Posture Rand. & \ding{51} & 37.27 $\pm$ 1.14 & 0.77 $\pm$ 0.01  & 0.77 $\pm$ 0.01 & 0.15 $\pm$ 0.00  & 234.46 $\pm$ 1.00 & 0.90$\pm$ 4.98e-4  & 0.72 $\pm$ 2.04e-4 \\

\ours w/ Hard Symmetry & \ding{51} & 75.53 $\pm$ 0.25 & 0.60 $\pm$ 0.00  & \textbf{0.31 $\pm$ 0.00} & 0.09 $\pm$ 0.00  & 84.95 $\pm$ 0.33 & 0.95 $\pm$ 3.12e-5  & 0.76 $\pm$ 2.49e-5 \\
\ours & \ding{51} & \textbf{94.40 $\pm$ 0.21} & \textbf{0.99 $\pm$0.00} & \textbf{0.31 $\pm$ 0.00} & \textbf{0.06 $\pm$ 0.00}  & \textbf{57.08 $\pm$ 0.20} & \textbf{0.95 $\pm$ 1.51e-4} & \textbf{0.76 $\pm$ 2.48e-5} \\

\midrule[0.6pt]
\multicolumn{9}{l}{\ding{184} \textit{\textbf{Getting Up from Prone Poses}}}\\
\midrule[0.6pt]
\citet{Learning2GetUp22}$^\dagger$ & \ding{55} & \textbf{98.99 $\pm$ 0.20} & \textbf{1.26 $\pm$ 0.00}  & 11.73 $\pm$ 0.01 & 0.76 $\pm$ 0.00  & 1015.27 $\pm$ 0.65 & 0.67 $\pm$ 2.24e-4  & 0.68 $\pm$ 6.41e-5 \\

\ours w/o Stage II & \ding{55} & 27.59 $\pm$ 0.28 & 0.82 $\pm$ 0.00  & 5.56 $\pm$ 0.36 & 0.45 $\pm$ 0.04  & 1213.07 $\pm$ 5.56 & 0.67 $\pm$ 4.71e-3  & 0.71 $\pm$ 2.17e-3 \\

\ours w/o Full URDF & \ding{55} & 89.59 $\pm$ 0.29 & 1.23 $\pm$ 0.00 & 0.44 $\pm$ 0.01 & 0.08 $\pm$ 0.00  & 77.61 $\pm$ 0.86 & 0.92 $\pm$ 2.88e-5  & 0.75 $\pm$ 3.19e-5 \\

\ours w/o Posture Rand. & \ding{51} & 30.25 $\pm$ 0.24 & 0.87 $\pm$ 0.02  & 1.05 $\pm$ 0.01 & 0.15 $\pm$ 0.00  & 208.23 $\pm$ 1.27 & 0.90 $\pm$ 3.06e-4  & 0.73 $\pm$ 1.01e-4 \\

\ours w/ Hard Symmetry & \ding{51}& 67.12 $\pm$ 0.34 & 1.09 $\pm$ 0.01  & 0.94  $\pm$ 0.01 & 0.23 $\pm$ 0.01  & 196.17 $\pm$ 3.68 & 0.91 $\pm$ 3.54e-5  & 0.76 $\pm$ 4.45e-5 \\

\ours & \ding{51} & 92.10 $\pm$ 0.46 & 1.24 $\pm$ 0.00  & \textbf{0.39 $\pm$ 0.01} & \textbf{0.07 $\pm$ 0.00}  & \textbf{69.98 $\pm$ 0.45} & \textbf{0.94 $\pm$ 1.82e-4}  & \textbf{0.77 $\pm$ 3.70e-4}  \\
\bottomrule[0.95pt] 
\end{tabular}
}
\end{table*}
\section{Implementation Details}\label{sec:exp_setup}

\subsection{Platform Configurations}
We use the Unitree G1 platform~\cite{UnitreeG124} in all real-world and simulation experiments. G1 is a medium-sized humanoid robot with 29 actuatable degrees of freedom (DoF) in total. Specifically, the upper body has 14 DoFs, the lower body has 12 DoFs, and the waist has 3 DoFs. As getting up does not involve object manipulation, we disable the 3 DOFs in the wrists, resulting in 23 DoFs in total.
Unlike previous robots, G1 has waist \texttt{yaw} and \texttt{roll} DoFs, and we find them useful for our getting-up task.
The robot has an IMU sensor for \texttt{roll} and \texttt{pitch} states, and the joint states can be obtained from the motor encoders. We use position control where the torque is derived by a PD controller operating at 50 Hz.

\subsection{Simulation Configurations}
We use Isaac Gym~\cite{IsaacGym21} for simulated training and evaluation.
We use a URDF with simplified collision for stage-I training and the official whole-body URDF from Unitree~\cite{UnitreeG124} for stage-II.
To accurately model the numerous contacts between the humanoid and the ground, we use a high simulation frequency of 1000 Hz, while the low-level PD controller frequency operates at 50 Hz.
More details can be found in \cref{app:training_details}.

\section{Simulation Results}\label{sec:sim_exp}

\subsection{Tasks}
We evaluate three tasks involved in the humanoid getting-up process:
\ding{182} \textit{getting up from supine poses}, \ding{183} \textit{rolling over from prone to supine poses}, and \ding{184} \textit{getting up from prone poses} which can be addressed by solving task \ding{183} and task \ding{182} consecutively.  Simulation tests are conducted with the full URDF.

\subsection{Baselines} We compare to the following baselines,
\begin{itemize}
    \item[a)] \textbf{RL with Simple Task Rewards (\citet{Learning2GetUp22}):} 
    This policy is trained with RL using rewards from~\citet{Learning2GetUp22} originally designed for physically animated characters instead of humanoid robots.
    Similar to our method, this baseline applies a three-stage strong-to-weak torque limit and motion speed curriculum for getting-up policy learning.
    Because \cite{Learning2GetUp22} does not consider sim2real deployment regularization and requirements (\eg, smoothness and collision mesh usage), policies learned through their scheme aren't appropriate for real-world humanoid deployment.
    \item[b)] \textbf{\ours w/o Stage II:} 
    Our policy trained with only stage I, where no deployment constraints are applied.
    \item[c)] \textbf{\ours w/o Full URDF:} 
    Our policy trained with two stages, but stage II uses the simplified collision mesh.
    \item[d)] \textbf{\ours w/o Posture Randomization:} 
    Our policy trained on a single canonical lying posture without any randomization of initialization postures.
    \item[e)] \textbf{\ours w/ Hard Symmetry:}
    Our policy trained using a humanoid with a symmetric controller.
    This symmetric controller follows the symmetry control principle of the manufacturer-provided controller baseline described in real-world experiments, which leads to bilaterally symmetric control.
    We set all \texttt{pitch} DoFs actions to be the same between the left and the right DoFs, while flipping the directions of all the \texttt{roll} and \texttt{yaw} actions.
    \item[f)] \textbf{\ours w/o Two-Stage Learning:}
    Our policy trained in a single stage with the full collision mesh, posture randomization, and all rewards and regularization terms applied at the same time.
\end{itemize}

\subsection{Metrics}\label{sec:metrics}
\noindent \textbullet~\textbf{Task Success.} 
i) Task success rate \textit{Success (\%)}:
For the getting-up task, the robot's head height must be $\geq$ 1.1m at termination, thus, the robot needs to continue to stand for success.
For the rolling-over task, the cosine between the robot's base, knee, and torso orientation and the target orientation when lying face up should be $\geq 0.9$, thus, the robot needs to move until lying face up.
ii) \textit{Task Metrics}: the head height (m) for the getting-up task, and the cosine of the angle between the robot's torso X-axis (sticking out to
the front from the torso) and the gravity, for the rolling-over task.

\noindent \textbullet~\textbf{Smoothness.} We measure the \textit{Action Jitter ($\text{rad}/\text{s}^3$)}, \textit{DoF Pos Jitter ($\text{rad}/\text{s}^3$)}, and mean \textit{Energy ($N\cdot m \cdot \text{rad} / \text{s}$ )} for action smoothness evaluation~\cite{EnergyEmergenceGaits21}. The jitter metrics are computed as the third derivative values~\cite{JitterMeasurement85}, which indicate the coordination stability of body movements.

\noindent \textbullet~\textbf{Safety.} We introduce safety scores $\mathcal{S}^{\text{Torque}}_{\delta,\alpha} \in [0, 1]$ and $\mathcal{S}^{\text{DoF}}_{\delta,\beta}\in[0,1]$ that measure the relative magnitude of commanded torque / DoF compared to the torque and DoF limits, where $\delta$ is a safety threshold. This is essential for robotic safety during execution, as large torques or DoFs will lead to overheating issues and cause mechanical and motor damage. Formally, these scores are defined as:
\[
\begin{aligned}
\mathcal{S}^{\text{Torque}}_{\delta, \alpha} &= 1 - \Bigg( \frac{\alpha}{T J} \sum_{t,j} \frac{| \tau_{t, j} |}{\tau_j^{\max}} + \frac{1-\alpha}{TJ} \sum_{t,j} \mathbbm{1} \Big( \frac{| \tau_{t, j} |}{\tau_j^{\max}} > \delta \Big) \Bigg),
\end{aligned}
\]
\[
\mathcal{S}^{\text{DoF}}_{\delta,\beta} = 1 - \Bigg( \frac{\beta}{T J} \sum_{t,j} \frac{| q_{t, j} |}{q_j^{\max}} + \frac{1 - \beta}{TJ} \sum_{t,j} \mathbbm{1} \Big( \frac{| q_{t, j} |}{q_j^{\max}} > \delta \Big) \Bigg),
\]
where $\tau_{t, j}$ and $q_{t, j}$ denote the applied torque and joint displacement at time step $t$ for joint $j$, respectively. $\tau_j^{\max}$ and $q_j^{\max}$ represent their respective limits, $T$ is the total number of time steps, and $J$ is the total number of joints. The threshold $\delta$ determines when a torque or displacement is considered excessive. The indicator function $\mathbbm{1}(\cdot)$ returns 1 if the condition is met and 0 otherwise. The parameters $\alpha, \beta \in [0,1]$ control the trade-off between peak and prolonged violations, ensuring a balanced assessment of safety risks.
In this paper, we use $\delta=0.8$, $\alpha=0.5$, $\beta=0.5$ as default during evaluation.

\subsection{Results and Analysis}\label{sec:sim_res}
\cref{tab:simulation_results} presents results based on policies tested on \texttt{held-out} subsets of our curated initial posture set \(\mathcal{P}\), \ie 10K val samples each from \(\mathcal{P}_{\text{supine}}\) and \(\mathcal{P}_{\text{prone}}\). \cref{fig:learning_curve} shows the learning curve for the getting-up task, where the termination base height reflects the robot’s ability to lift its body, and body uprightness indicates whether it achieves a stable standing posture.

\begin{figure*}
  \centering
    \includegraphics[width=\linewidth]{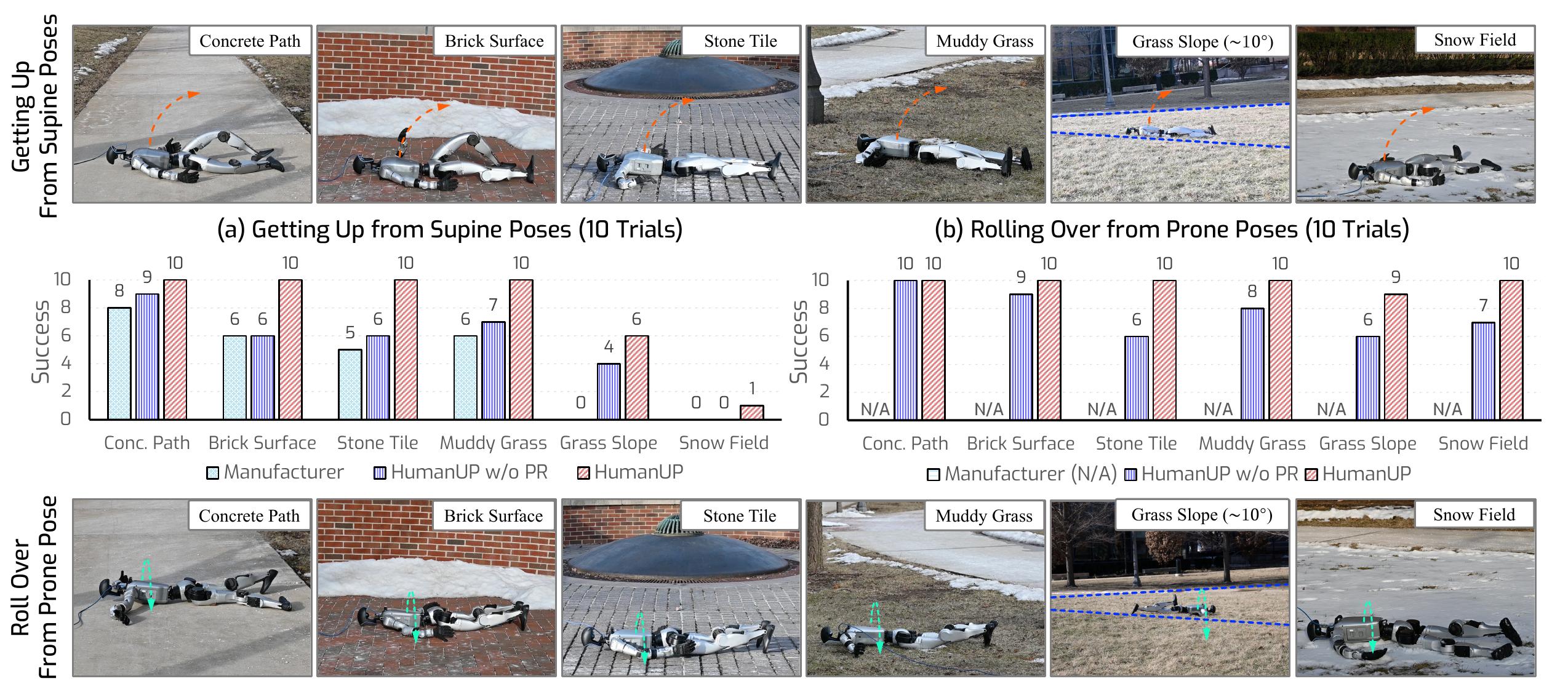}
  \caption{\textbf{Real-world results}. We evaluate \ours (ours) in several real setups that span diverse surface properties, including both man-made and natural surfaces, and cover a wide range of roughness (rough concrete to slippery snow), bumpiness (flat concrete to tiles), ground compliance (completely firm concrete to being swampy muddy grass), and slope (flat to about $10^\circ$). We compare \ours with G1's manufacturer-provided controller and \ours w/o posture randomization (PR). \ours succeeds more consistently (78.3\% vs 41.7\%) and can solve terrains that the manufacturer-provided controller can't.
  }\label{fig:realworld_res_main}
\end{figure*}

\subsubsection{Ignoring Torque / Control Limits Leads to Undeployable Policies}
While \cite{Learning2GetUp22} and \ours achieve similar success rates, the smoothness and safety metrics for \cite{Learning2GetUp22} are significantly worse than \ours.
For example, the average action jitter metric is nearly 18$\times$ higher than \ours. Actions from \cite{Learning2GetUp22} are highly unstable and unsafe and thus cannot be safely deployed to the real robot. Furthermore, \cite{Learning2GetUp22} learns a very fast getting-up motion that keeps jumping after getting up. See visualization \cite{Learning2GetUp22}'s getting up motion in \cref{app:baseline_result}. A similar trend can be seen when comparing \ours to \ours w/o \stwo. While \ours w/o \stwo also solves the task to some extent, it achieves unsatisfying smoothness and safety metrics similar making it inappropriate for real-world deployment. Thus, the regularization imposed in \stwo is essential to making policies more amenable to Sim2Real transfer.

\subsubsection{Importance of Learning via a Curriculum}
\begin{figure}[ht!]
  \centering
    \includegraphics[width=\linewidth]{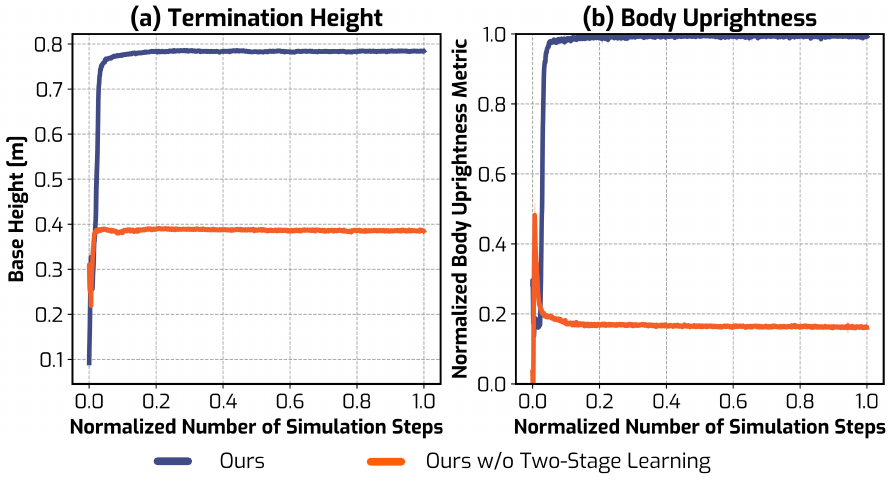}
  \caption{\textbf{Learning curve}. 
  (a) Termination height of the torso, indicating whether the robot can lift the body.
  (b) Body uprightness, computed as the projected gravity on the \(z\)-axis, normalized to \([0,1]\) for better comparison.
  The overall number of simulation sampling steps is about 5B, normalized to $[0,1]$.
  }\label{fig:learning_curve}
  \vspace{-0.5cm}
\end{figure}

So, while it is clear that we need to incorporate strong control regularization for good safety metrics and Sim2Real transfer, our 2 stage process is better than doing it in a single stage. In fact, as plotted in \cref{fig:learning_curve}, \ours w/o Two Stage Learning where the policy is trained in a single stage using all sim2real regularization fails to solve the task. This is because the strict Sim2Real regularization makes task learning extremely challenging. Our two-stage curriculum successfully incorporates both aspects: it learns to solve the task, but the policy also operates safely.

\subsubsection{Full URDF vs. Simplified URDF}
Somewhat surprisingly, even though \ours w/o Full URDF was trained without the full URDF mesh, it generalizes fine when tested with the full URDF in simulation, as reported in \cref{tab:simulation_results}. However, we found poor transfer of this policy to the real world. 
It failed on all 5 trials on a simple flat terrain. We believe the poor real-world performance was because of the mismatch between the contact it was expecting and the contact that actually happened.

\subsubsection{Posture randomization helps}
\ours w/o posture randomization (PR) works much worse than \ours, suggesting that PR is necessary for generalizable control.

\subsubsection{Soft symmetry vs. hard symmetry}
Compared to \ours w/ Hard Symmetry, \ours achieves better task success in \cref{tab:simulation_results}, particularly for the rolling-over task, which is very difficult with symmetric commands.

\begin{figure*}
  \centering
    \includegraphics[width=\linewidth]{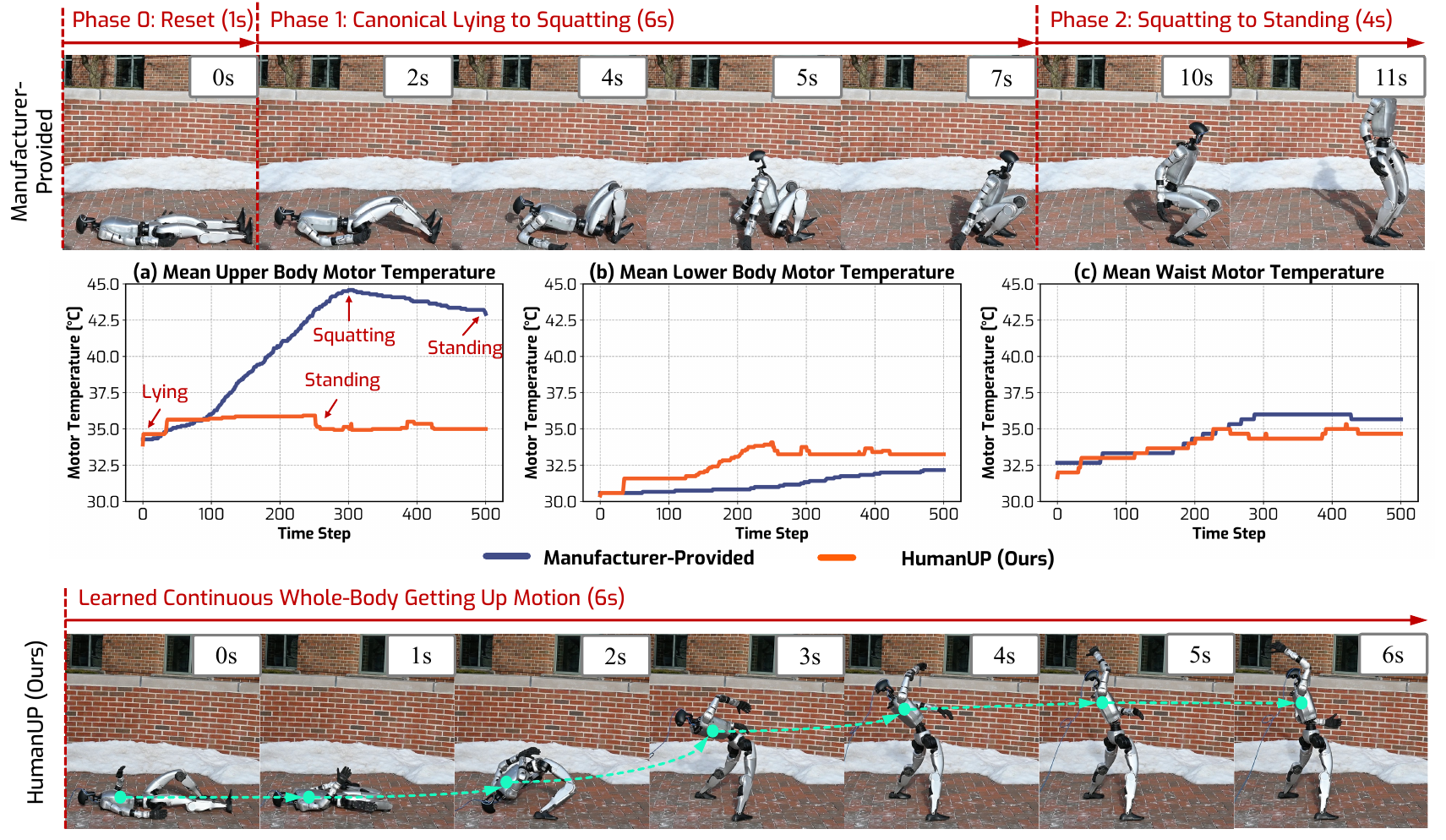}
  \caption{\textbf{Getting up execution comparison with G1's manufacturer-provided controller}. 
  The manufacturer-provided controller uses a handcrafted motion trajectory, which can be divided into three phases, while our \ours learns a continuous and more efficient whole-body getting-up motion.
  Our \ours enables the humanoid to get up within 6 seconds, half of the manufacturer-provided controller's 11 seconds of control.
  (a), (b), and (c) record the corresponding mean motor temperature of the upper body, lower body, and waist, respectively. G1's manufacturer-provided controller's execution causes the arm motors to heat up significantly, whereas our policy makes more use of the leg motors that are stronger (higher torque limit of 83N as opposed to 25N for the arm motors) and thus able to take more load.
  }\label{fig:compare_g1_controller}
\end{figure*}
\begin{figure*}
  \centering
    \includegraphics[width=\linewidth]{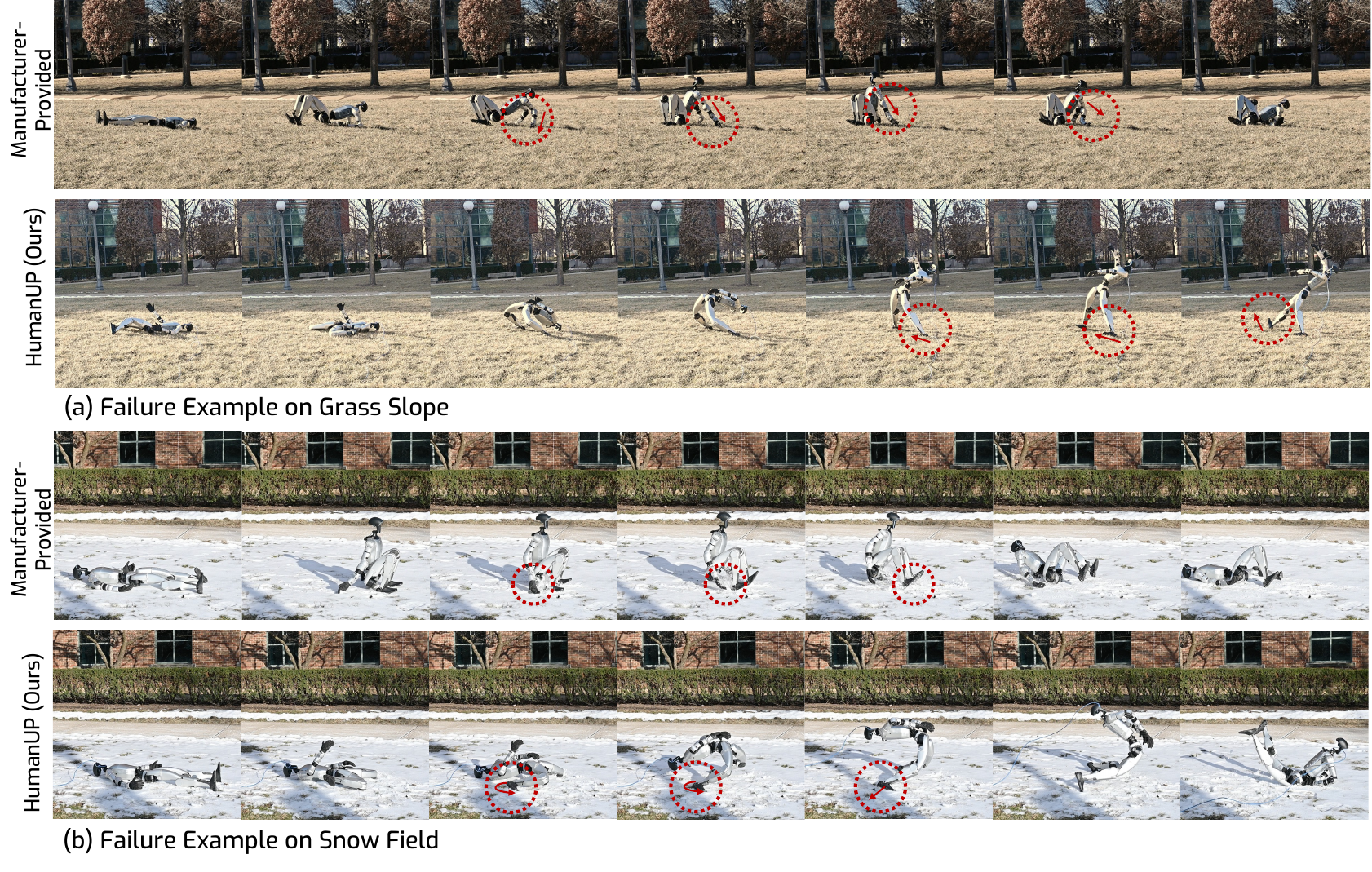}
  \caption{\textbf{Qualitative examples of failure modes on grass slope and snow field.} G1’s manufacturer-provided controller isn't able to squat on the sloping grass and slips on the slope. \ours policy can partially get up on both the slope and the snow, but falls due to unstable foot placement on the slope and slippage on the snow.
  }\label{fig:failure_examples}
\end{figure*}
\section{Real World Results}\label{sec:real_exp}
We also tested \ours policies in the real world on G1 robot. 
Our real-world test bed consists of 6 different terrains shown in \cref{fig:realworld_res_main}: concrete, brick, stone tiles, muddy grass, grassy slope, and a snow field. 
These terrains span diverse surface properties, including both man-made and natural surfaces, and cover a wide range of roughness (rough concrete to slippery snow), bumpiness (flat concrete to tiles), ground compliance (completely firm concrete to being swampy muddy grasp), and slope (flat to $\sim 10^\circ$). 
We tested two tasks: a) getting up from supine poses, and b) rolling over from prone to supine poses. 

We compare our policy with
1) \textbf{Manufacturer-provided Controller} and 2) a high-performing ablation of \ours (\textbf{\ours w/o posture randomization}). 
The manufacturer-provided controller, which comes with the robot G1, tracks a hand-crafted trajectory in three phases shown in \cref{fig:compare_g1_controller}:
Phase 0 brings the robot to a canonical lying pose; Phase 1 first props up and then slides the torso forward using hands, followed by bending legs to squat; Phase 2 uses its waist to tilt up the torso to stand up from squatting.
Motions in phase 1 and phase 2 are \textit{symmetric}, and this controller only works for supine poses.

\subsection{Results}\label{sec:real_res}
\cref{fig:realworld_res_main} presents experimental results. 
Overall, we find that \ours policies perform better than the manufacturer-provided controller and \ours without posture randomization (PR). We discuss the results and observed behavior further.

\subsubsection{Getting up from supine poses} 
The manufacturer-provided controller works under nominal conditions, \ie, firm, flat concrete ground with a reasonable friction value. 
However, it starts to fail on more challenging terrains. 
For the bumpier and rougher terrains (brick surface and stone tiles), the arms may get stuck between bumps, causing failures. 
On slopes, the robot fails to squat or hoist itself up due to both the resistance of the grassland and the unstable squatting pose prone to falling caused by slopes.
On the compliant ground, the robot gets destabilized.
On slippery snow, the robot slips. 

Both versions of \ours outperform the manufacturer-provided controller. Trained with terrain and domain randomization, they are robust to real-world variations such as slipperiness, bumps, and slopes. Dynamics randomization further enhances resilience to minor perturbations like slippage or ground compliance. The full method, incorporating posture randomization, performs better than the variant without it, as it is specifically trained to handle diverse initial configurations. Overall, \ours achieves a 78.3\% getting-up success rate.

\subsubsection{Rolling over from prone to supine poses} Findings are similar for the rolling over task. As noted, the manufacturer-provided controller can't handle this situation. The full model exhibits more robust performance than the model trained without posture randomization. Rolling over seems to be easier than getting up, \ours achieves a 98.3\% success rate. 

\subsection{Motion analysis}
\cref{fig:compare_g1_controller} shows the motion and motor temperatures for the manufacturer-provided controller and \ours policy.

\subsubsection{Motor temperature}
The manufacturer-provided controller uses the arms during Phase 1 of getting up.
\cref{fig:compare_g1_controller}\,(a) shows that the default controller's execution causes the arm motors to heat up significantly more when compared to \ours execution. 
Our policy makes more use of the leg motors that are stronger (higher torque limit of $83N$ as opposed to $25N$ for the arm motors) and thus able to take more load. 

\subsubsection{Efficiency}
\ours gets the robot to stand successfully within about 6 seconds through a smooth and continuous motion, which is over 2$\times$ more efficient than the manufacturer-provided controller, which takes nearly 11 seconds.

\subsection{Failure Mode Analysis} 
\cref{fig:failure_examples} shows example failure modes for the manufacturer-provided controller and \ours on challenging terrains.
\cref{fig:failure_examples}\,(a) shows that the manufacturer-provided controller tries to utilize the robot's hands to squat, while the sloping ground prevents it from getting to the full squatting pose due to high friction and weak waist torques to move against the dumping tendency.
In contrast, \ours manages to lift the body, while the sloping ground sometimes causes an unstable foot orientation.
\cref{fig:failure_examples}\,(b) shows that on even more challenging terrains like snow fields, both manufacturer-provided and \ours controllers may fail due to the slippery and deformable ground.
\section{Limitations}
\ours has several limitations:
1) Motions discovered in \sone could be incompatible with stronger control regularization used in \stwo. We didn't encounter this issue in our experiments, possibly because of the weak control regularization used in \sone and the use of $8\times$ slower motion in \stwo.
2) \ours depends on high-performance physics simulators (IsaacGym~\cite{IsaacGym21}) running at high frequency (\eg, 1 kHz). Simulation speed and fidelity for more complex tasks involving perception and contacts remain a challenge.
Recent advances such as Genesis~\cite{Genesis24}, Mujoco Playground~\cite{MujocoPlayground25}, and Roboverse~\cite{Roboverse25} could help address these limitations.
3) The RL formulation in \ours is under-specified~\cite{TaskSpecificationProblem22} and may lead to reward hacking~\cite{RewardHackingBlog24}, complicating precise alignment with natural human behaviors. For instance, our learned motions sometimes include unnatural hand raising for balance.
4) Extending \ours to handle more complex terrains like stairs or uneven surfaces remains under-explored, while humanoid robots may fall more easily on such terrains. Encouraging adaptive behaviors involving strong-arm usage on more powerful platforms may be useful to properly handle such situations.

\section{Discussion}
In this paper, we tackle the problem of {\it learning} getting-up controllers for real-world humanoid robots. Different from locomotion tasks, getting up involves complex contact patterns that are not known apriori. 
We develop a two-stage solution for this problem based on reinforcement learning and sim-to-real. Stage I finds a solution under minimal constraints, while \stwo learns to track the trajectory discovered in \sone under regularization on control and from varied starting poses and on varied terrains. We found this two stage strategy to be effective both in simulation and the real world. Specifically, it enabled us to get a real-world G1 humanoid to stand up from a supine pose and roll over from a supine pose to a prone pose on different terrains and from different starting poses. \ours achieves a higher success rate than G1's manufacturer-provided controller and expands the capabilities of the G1 robot.

We hope our learned policies for automatic fall recovery will be useful to researchers and practitioners, while our two-stage learning framework may be helpful for other problems that require figuring out complex contact patterns.


\section*{Acknowledgments}
This material is based upon work supported by an NSF CAREER Award (IIS2143873) and a DURIP grant (N00014-23-1-2166). We also thank the Coordinated Science Laboratory for providing experimental space.

{
\bibliographystyle{plainnat}
\bibliography{main}

\begin{thebibliography}{83}
\providecommand{\natexlab}[1]{#1}
\providecommand{\url}[1]{\texttt{#1}}
\expandafter\ifx\csname urlstyle\endcsname\relax
  \providecommand{\doi}[1]{doi: #1}\else
  \providecommand{\doi}{doi: \begingroup \urlstyle{rm}\Url}\fi

\bibitem[Boo()]{Booster}
Booster robotics.
\newblock URL \url{https://www.boosterobotics.com/}.

\bibitem[Adu-Bredu et~al.(2023)Adu-Bredu, Gibson, and Grizzle]{KinodynamicFabrics23}
Alphonsus Adu-Bredu, Grant Gibson, and Jessy Grizzle.
\newblock Exploring kinodynamic fabrics for reactive whole-body control of underactuated humanoid robots.
\newblock In \emph{2023 IEEE/RSJ International Conference on Intelligent Robots and Systems (IROS)}, pages 10397--10404. IEEE, 2023.

\bibitem[Agrawal(2022)]{TaskSpecificationProblem22}
Pulkit Agrawal.
\newblock The task specification problem.
\newblock In \emph{Conference on Robot Learning}, pages 1745--1751. PMLR, 2022.

\bibitem[Ahn(2023)]{bookwalkingrunning2023}
Min~Sung Ahn.
\newblock \emph{Development and Real-Time Optimization-based Control of a Full-sized Humanoid for Dynamic Walking and Running}.
\newblock University of California, Los Angeles, 2023.

\bibitem[Anonymous(2025)]{HierachicalWorldModel25}
Anonymous.
\newblock Hierarchical world models as visual whole-body humanoid controllers.
\newblock In \emph{The Thirteenth International Conference on Learning Representations}, 2025.

\bibitem[Authors(2024)]{Genesis24}
Genesis Authors.
\newblock Genesis: A universal and generative physics engine for robotics and beyond, December 2024.
\newblock URL \url{https://github.com/Genesis-Embodied-AI/Genesis}.

\bibitem[Bengio et~al.(2009)Bengio, Louradour, Collobert, and Weston]{CurriculumLearning09}
Yoshua Bengio, J\'{e}r\^{o}me Louradour, Ronan Collobert, and Jason Weston.
\newblock Curriculum learning.
\newblock In \emph{Proceedings of the 26th Annual International Conference on Machine Learning}, ICML '09, page 41–48, New York, NY, USA, 2009. Association for Computing Machinery.

\bibitem[Chen et~al.(2024)Chen, He, Wang, Liao, Ze, Li, Sastry, Wu, Sreenath, Gupta, et~al.]{LCP24}
Zixuan Chen, Xialin He, Yen-Jen Wang, Qiayuan Liao, Yanjie Ze, Zhongyu Li, S~Shankar Sastry, Jiajun Wu, Koushil Sreenath, Saurabh Gupta, et~al.
\newblock Learning smooth humanoid locomotion through lipschitz-constrained policies.
\newblock \emph{arXiv preprint arXiv:2410.11825}, 2024.

\bibitem[Cheng et~al.(2024{\natexlab{a}})Cheng, Ji, Chen, Yang, Yang, and Wang]{Exbody24}
Xuxin Cheng, Yandong Ji, Junming Chen, Ruihan Yang, Ge~Yang, and Xiaolong Wang.
\newblock {Expressive Whole-Body Control for Humanoid Robots}.
\newblock In \emph{Proceedings of Robotics: Science and Systems}, Delft, Netherlands, July 2024{\natexlab{a}}.

\bibitem[Cheng et~al.(2024{\natexlab{b}})Cheng, Shi, Agarwal, and Pathak]{ExtremeParkour24}
Xuxin Cheng, Kexin Shi, Ananye Agarwal, and Deepak Pathak.
\newblock Extreme parkour with legged robots.
\newblock In \emph{{IEEE} International Conference on Robotics and Automation, {ICRA} 2024, Yokohama, Japan, May 13-17, 2024}, pages 11443--11450. {IEEE}, 2024{\natexlab{b}}.

\bibitem[Chentanez et~al.(2018)Chentanez, M{\"{u}}ller, Macklin, Makoviychuk, and Jeschke]{PhysicsBasedMocapImitationDRL18}
Nuttapong Chentanez, Matthias M{\"{u}}ller, Miles Macklin, Viktor Makoviychuk, and Stefan Jeschke.
\newblock Physics-based motion capture imitation with deep reinforcement learning.
\newblock In Panayiotis Charalambous, Yiorgos Chrysanthou, Ben Jones, and Jehee Lee, editors, \emph{Proceedings of the 11th Annual International Conference on Motion, Interaction, and Games, {MIG} 2018, Limassol, Cyprus, November 08-10, 2018}, pages 1:1--1:10. {ACM}, 2018.

\bibitem[Chignoli et~al.(2021)Chignoli, Kim, Stanger-Jones, and Kim]{MITHumanoid21}
Matthew Chignoli, Donghyun Kim, Elijah Stanger-Jones, and Sangbae Kim.
\newblock The mit humanoid robot: Design, motion planning, and control for acrobatic behaviors.
\newblock In \emph{2020 IEEE-RAS 20th International Conference on Humanoid Robots (Humanoids)}, pages 1--8, 2021.

\bibitem[Clark and Amodei(2016)]{RewardHackingBlog24}
Jack Clark and Dario Amodei.
\newblock Faulty reward functions in the wild.
\newblock 2016.
\newblock URL \url{https://openai.com/index/faulty-reward-functions/}.

\bibitem[Crowley et~al.(2023)Crowley, Dao, Duan, Green, Hurst, and Fern]{BipedalRunning23}
Devin Crowley, Jeremy Dao, Helei Duan, Kevin Green, Jonathan Hurst, and Alan Fern.
\newblock Optimizing bipedal locomotion for the 100m dash with comparison to human running.
\newblock In \emph{2023 IEEE International Conference on Robotics and Automation (ICRA)}, pages 12205--12211, 2023.

\bibitem[Di~Carlo et~al.(2018)Di~Carlo, Wensing, Katz, Bledt, and Kim]{DynamicLocomotionMITConvexMPC18}
Jared Di~Carlo, Patrick~M. Wensing, Benjamin Katz, Gerardo Bledt, and Sangbae Kim.
\newblock Dynamic locomotion in the mit cheetah 3 through convex model-predictive control.
\newblock In \emph{2018 IEEE/RSJ International Conference on Intelligent Robots and Systems (IROS)}, pages 1--9, 2018.

\bibitem[Flash and Hogan(1985)]{JitterMeasurement85}
Tamar Flash and Neville Hogan.
\newblock The coordination of arm movements: an experimentally confirmed mathematical model.
\newblock \emph{Journal of neuroscience}, 5\penalty0 (7):\penalty0 1688--1703, 1985.

\bibitem[Frezzato et~al.(2022)Frezzato, Tangri, and Andrews]{frezzato2022synthesizing}
Anthony Frezzato, Arsh Tangri, and Sheldon Andrews.
\newblock Synthesizing get-up motions for physics-based characters.
\newblock In \emph{Computer Graphics Forum}, volume~41, pages 207--218. Wiley Online Library, 2022.

\bibitem[Fu et~al.(2021)Fu, Kumar, Malik, and Pathak]{EnergyEmergenceGaits21}
Zipeng Fu, Ashish Kumar, Jitendra Malik, and Deepak Pathak.
\newblock Minimizing energy consumption leads to the emergence of gaits in legged robots.
\newblock In Aleksandra Faust, David Hsu, and Gerhard Neumann, editors, \emph{Conference on Robot Learning, 8-11 November 2021, London, {UK}}, volume 164 of \emph{Proceedings of Machine Learning Research}, pages 928--937. {PMLR}, 2021.

\bibitem[Fu et~al.(2022)Fu, Cheng, and Pathak]{DeepWholeBodyControl22}
Zipeng Fu, Xuxin Cheng, and Deepak Pathak.
\newblock Deep whole-body control: Learning a unified policy for manipulation and locomotion.
\newblock In Karen Liu and Dana~Kulic andD Jeffrey~Ichnowski, editors, \emph{Conference on Robot Learning, CoRL 2022, 14-18 December 2022, Auckland, New Zealand}, volume 205 of \emph{Proceedings of Machine Learning Research}, pages 138--149. {PMLR}, 2022.

\bibitem[Fu et~al.(2024)Fu, Zhao, Wu, Wetzstein, and Finn]{HumanPlus24}
Zipeng Fu, Qingqing Zhao, Qi~Wu, Gordon Wetzstein, and Chelsea Finn.
\newblock Humanplus: Humanoid shadowing and imitation from humans.
\newblock In \emph{8th Annual Conference on Robot Learning}, 2024.

\bibitem[Fujiwara et~al.(2002)Fujiwara, Kanehiro, Kajita, Kaneko, Yokoi, and Hirukawa]{UKEMI02}
Kiyoshi Fujiwara, Fumio Kanehiro, Shuuji Kajita, Kenji Kaneko, Kazuhito Yokoi, and Hirohisa Hirukawa.
\newblock Ukemi: Falling motion control to minimize damage to biped humanoid robot.
\newblock In \emph{IEEE/RSJ international conference on Intelligent robots and systems}, volume~3, pages 2521--2526. IEEE, 2002.

\bibitem[Galliker et~al.(2022)Galliker, Csomay-Shanklin, Grandia, Taylor, Farshidian, Hutter, and Ames]{galliker2022planar}
Manuel~Y Galliker, Noel Csomay-Shanklin, Ruben Grandia, Andrew~J Taylor, Farbod Farshidian, Marco Hutter, and Aaron~D Ames.
\newblock Planar bipedal locomotion with nonlinear model predictive control: Online gait generation using whole-body dynamics.
\newblock In \emph{2022 IEEE-RAS 21st International Conference on Humanoid Robots (Humanoids)}, pages 622--629. IEEE, 2022.

\bibitem[Gao et~al.(2024)Gao, Wang, Xiao, Wang, Wang, Cao, Hu, Liu, Dai, and Pang]{CooHOI24}
Jiawei Gao, Ziqin Wang, Zeqi Xiao, Jingbo Wang, Tai Wang, Jinkun Cao, Xiaolin Hu, Si~Liu, Jifeng Dai, and Jiangmiao Pang.
\newblock Coohoi: Learning cooperative human-object interaction with manipulated object dynamics.
\newblock \emph{arXiv preprint arXiv:2406.14558}, 2024.

\bibitem[Gaspard et~al.(2024)Gaspard, Duclusaud, Passault, Daniel, and Ly]{FRASA24}
Cl{\'e}ment Gaspard, Marc Duclusaud, Gr{\'e}goire Passault, M{\'e}lodie Daniel, and Olivier Ly.
\newblock Frasa: An end-to-end reinforcement learning agent for fall recovery and stand up of humanoid robots.
\newblock \emph{arXiv preprint arXiv:2410.08655}, 2024.

\bibitem[Geng et~al.(2025)Geng, Wang, Wei, Li, Wang, An, Cheng, Lou, Li, Wang, Liang, Goetting, Xu, Chen, Qian, Geng, Mao, Wan, Zhang, Lyu, Zhao, Zhang, Zhang, Zhao, Lu, Ding, Gong, Wang, Kuang, Wu, Jia, Sferrazza, Dong, Huang, Sreenath, Wang, Malik, and Abbeel]{Roboverse25}
Haoran Geng, Feishi Wang, Songlin Wei, Yuyang Li, Bangjun Wang, Boshi An, Charlie~Tianyue Cheng, Haozhe Lou, Peihao Li, Yen-Jen Wang, Yutong Liang, Dylan Goetting, Chaoyi Xu, Haozhe Chen, Yuxi Qian, Yiran Geng, Jiageng Mao, Weikang Wan, Mingtong Zhang, Jiangran Lyu, Siheng Zhao, Jiazhao Zhang, Jialiang Zhang, Chengyang Zhao, Haoran Lu, Yufei Ding, Ran Gong, Yuran Wang, Yuxuan Kuang, Ruihai Wu, Baoxiong Jia, Carlo Sferrazza, Hao Dong, Siyuan Huang, Koushil Sreenath, Yue Wang, Jitendra Malik, and Pieter Abbeel.
\newblock Roboverse: Towards a unified platform, dataset and benchmark for scalable and generalizable robot learning, April 2025.
\newblock URL \url{https://github.com/RoboVerseOrg/RoboVerse}.

\bibitem[Gonz{\'a}lez-Fierro et~al.(2013)Gonz{\'a}lez-Fierro, Balaguer, Swann, and Nanayakkara]{HumanoidStandingUpLfDMultimodalReward13}
Miguel Gonz{\'a}lez-Fierro, Carlos Balaguer, Nicola Swann, and Thrishantha Nanayakkara.
\newblock A humanoid robot standing up through learning from demonstration using a multimodal reward function.
\newblock In \emph{2013 13th IEEE-RAS International Conference on Humanoid Robots (Humanoids)}, pages 74--79. IEEE, 2013.

\bibitem[Grizzle et~al.(2009)Grizzle, Hurst, Morris, Park, and Sreenath]{MABEL09}
J.~W. Grizzle, Jonathan~W. Hurst, Benjamin Morris, Hae{-}Won Park, and Koushil Sreenath.
\newblock Mabel, a new robotic bipedal walker and runner.
\newblock In \emph{American Control Conference, {ACC} 2009. St. Louis, Missouri, USA, June 10-12, 2009}, pages 2030--2036. {IEEE}, 2009.

\bibitem[Gu et~al.(2024{\natexlab{a}})Gu, Wang, Zhu, Shi, Guo, Liu, and Chen]{DenoisingWorldModel24}
Xinyang Gu, Yen-Jen Wang, Xiang Zhu, Chengming Shi, Yanjiang Guo, Yichen Liu, and Jianyu Chen.
\newblock {Advancing Humanoid Locomotion: Mastering Challenging Terrains with Denoising World Model Learning}.
\newblock In \emph{Proceedings of Robotics: Science and Systems}, Delft, Netherlands, July 2024{\natexlab{a}}.

\bibitem[Gu et~al.(2024{\natexlab{b}})Gu, Wang, Zhu, Shi, Guo, Liu, and Chen]{advancinglocomotion2024}
Xinyang Gu, Yen-Jen Wang, Xiang Zhu, Chengming Shi, Yanjiang Guo, Yichen Liu, and Jianyu Chen.
\newblock Advancing humanoid locomotion: Mastering challenging terrains with denoising world model learning.
\newblock \emph{arXiv preprint arXiv:2408.14472}, 2024{\natexlab{b}}.

\bibitem[Gu et~al.(2025)Gu, Li, Shen, Yu, Xie, McCrory, Cheng, Shamsah, Griffin, Liu, et~al.]{gu2025humanoid}
Zhaoyuan Gu, Junheng Li, Wenlan Shen, Wenhao Yu, Zhaoming Xie, Stephen McCrory, Xianyi Cheng, Abdulaziz Shamsah, Robert Griffin, C~Karen Liu, et~al.
\newblock Humanoid locomotion and manipulation: Current progress and challenges in control, planning, and learning.
\newblock \emph{arXiv preprint arXiv:2501.02116}, 2025.

\bibitem[H{\"a}m{\"a}l{\"a}inen et~al.(2014)H{\"a}m{\"a}l{\"a}inen, Eriksson, Tanskanen, Kyrki, and Lehtinen]{OnlineMotionSynthesis14}
Perttu H{\"a}m{\"a}l{\"a}inen, Sebastian Eriksson, Esa Tanskanen, Ville Kyrki, and Jaakko Lehtinen.
\newblock Online motion synthesis using sequential monte carlo.
\newblock \emph{ACM Transactions on Graphics (TOG)}, 33\penalty0 (4):\penalty0 1--12, 2014.

\bibitem[He et~al.(2024{\natexlab{a}})He, Luo, He, Xiao, Zhang, Zhang, Kitani, Liu, and Shi]{OmniH2O24}
Tairan He, Zhengyi Luo, Xialin He, Wenli Xiao, Chong Zhang, Weinan Zhang, Kris~M. Kitani, Changliu Liu, and Guanya Shi.
\newblock Omnih2o: Universal and dexterous human-to-humanoid whole-body teleoperation and learning.
\newblock In \emph{8th Annual Conference on Robot Learning}, 2024{\natexlab{a}}.

\bibitem[He et~al.(2024{\natexlab{b}})He, Luo, Xiao, Zhang, Kitani, Liu, and Shi]{Human2HumanTeleoperation24}
Tairan He, Zhengyi Luo, Wenli Xiao, Chong Zhang, Kris Kitani, Changliu Liu, and Guanya Shi.
\newblock Learning human-to-humanoid real-time whole-body teleoperation.
\newblock In \emph{{IEEE/RSJ} International Conference on Intelligent Robots and Systems, {IROS} 2024, Abu Dhabi, United Arab Emirates, October 14-18, 2024}, pages 8944--8951. {IEEE}, 2024{\natexlab{b}}.

\bibitem[He et~al.(2024{\natexlab{c}})He, Xiao, Lin, Luo, Xu, Jiang, Kautz, Liu, Shi, Wang, et~al.]{Hover24}
Tairan He, Wenli Xiao, Toru Lin, Zhengyi Luo, Zhenjia Xu, Zhenyu Jiang, Jan Kautz, Changliu Liu, Guanya Shi, Xiaolong Wang, et~al.
\newblock Hover: Versatile neural whole-body controller for humanoid robots.
\newblock \emph{arXiv preprint arXiv:2410.21229}, 2024{\natexlab{c}}.

\bibitem[Hirai et~al.(1998)Hirai, Hirose, Haikawa, and Takenaka]{HondaHumanoid98}
Kazuo Hirai, Masato Hirose, Yuji Haikawa, and Toru Takenaka.
\newblock The development of honda humanoid robot.
\newblock In \emph{Proceedings. 1998 IEEE international conference on robotics and automation (Cat. No. 98CH36146)}, volume~2, pages 1321--1326. IEEE, 1998.

\bibitem[Inaba et~al.(1996)Inaba, Igarashi, Kagami, and Inoue]{Humanoid35DoF96}
Masayuki Inaba, Takashi Igarashi, Satoshi Kagami, and Hirochika Inoue.
\newblock A 35 dof humanoid that can coordinate arms and legs in standing up, reaching and grasping an object.
\newblock In \emph{Proceedings of IEEE/RSJ International Conference on Intelligent Robots and Systems. IROS'96}, volume~1, pages 29--36. IEEE, 1996.

\bibitem[Jeong and Lee(2016)]{StandUpSymmetry16}
Heejin Jeong and Daniel~D. Lee.
\newblock Efficient learning of stand-up motion for humanoid robots with bilateral symmetry.
\newblock In \emph{2016 {IEEE/RSJ} International Conference on Intelligent Robots and Systems, {IROS} 2016, Daejeon, South Korea, October 9-14, 2016}, pages 1544--1549. {IEEE}, 2016.

\bibitem[Ji et~al.(2024)Ji, Peng, Liu, Li, Yang, Cheng, and Wang]{Exbody2_24}
Mazeyu Ji, Xuanbin Peng, Fangchen Liu, Jialong Li, Ge~Yang, Xuxin Cheng, and Xiaolong Wang.
\newblock Exbody2: Advanced expressive humanoid whole-body control.
\newblock \emph{arXiv preprint arXiv:2412.13196}, 2024.

\bibitem[Ji et~al.(2023)Ji, Margolis, and Agrawal]{dribblebot2023}
Yandong Ji, Gabriel~B Margolis, and Pulkit Agrawal.
\newblock Dribblebot: Dynamic legged manipulation in the wild.
\newblock In \emph{2023 IEEE International Conference on Robotics and Automation (ICRA)}, pages 5155--5162. IEEE, 2023.

\bibitem[Kanehiro et~al.(2003)Kanehiro, Kaneko, Fujiwara, Harada, Kajita, Yokoi, Hirukawa, Akachi, and Isozumi]{FirstHumanoidGetUp03}
Fumio Kanehiro, Kenji Kaneko, Kiyoshi Fujiwara, Kensuke Harada, Shuuji Kajita, Kazuhito Yokoi, Hirohisa Hirukawa, Kazuhiko Akachi, and Takakatsu Isozumi.
\newblock The first humanoid robot that has the same size as a human and that can lie down and get up.
\newblock In \emph{2003 IEEE International Conference on Robotics and Automation (Cat. No. 03CH37422)}, volume~2, pages 1633--1639. IEEE, 2003.

\bibitem[Kanehiro et~al.(2007)Kanehiro, Fujiwara, Hirukawa, Nakaoka, and Morisawa]{GettingUpMotionPlanning07}
Fumio Kanehiro, Kiyoshi Fujiwara, Hirohisa Hirukawa, Shin'ichiro Nakaoka, and Mitsuharu Morisawa.
\newblock Getting up motion planning using mahalanobis distance.
\newblock In \emph{Proceedings 2007 IEEE International Conference on Robotics and Automation}, pages 2540--2545. IEEE, 2007.

\bibitem[Kato(1973)]{WABOT1_73}
Ichiro Kato.
\newblock Development of wabot 1.
\newblock \emph{Biomechanism}, 2:\penalty0 173--214, 1973.

\bibitem[Kavraki et~al.(1996)Kavraki, Svestka, Latombe, and Overmars]{StateTransitionGraph96}
Lydia~E. Kavraki, Petr Svestka, Jean{-}Claude Latombe, and Mark~H. Overmars.
\newblock Probabilistic roadmaps for path planning in high-dimensional configuration spaces.
\newblock \emph{{IEEE} Trans. Robotics Autom.}, 12\penalty0 (4):\penalty0 566--580, 1996.

\bibitem[Krotkov et~al.(2018)Krotkov, Hackett, Jackel, Perschbacher, Pippine, Strauss, Pratt, and Orlowski]{krotkov2018darpa}
Eric Krotkov, Douglas Hackett, Larry Jackel, Michael Perschbacher, James Pippine, Jesse Strauss, Gill Pratt, and Christopher Orlowski.
\newblock The darpa robotics challenge finals: Results and perspectives.
\newblock \emph{The DARPA robotics challenge finals: Humanoid robots to the rescue}, pages 1--26, 2018.

\bibitem[Kuindersma et~al.(2016)Kuindersma, Deits, Fallon, Valenzuela, Dai, Permenter, Koolen, Marion, and Tedrake]{OptimizationBasedLocomotion16}
Scott Kuindersma, Robin Deits, Maurice Fallon, Andr{\'e}s Valenzuela, Hongkai Dai, Frank Permenter, Twan Koolen, Pat Marion, and Russ Tedrake.
\newblock Optimization-based locomotion planning, estimation, and control design for the atlas humanoid robot.
\newblock \emph{Autonomous robots}, 40:\penalty0 429--455, 2016.

\bibitem[Kumar et~al.(2021)Kumar, Fu, Pathak, and Malik]{RMA21}
Ashish Kumar, Zipeng Fu, Deepak Pathak, and Jitendra Malik.
\newblock {RMA:} rapid motor adaptation for legged robots.
\newblock In \emph{Robotics: Science and Systems XVII, Virtual Event, July 12-16, 2021}, 2021.

\bibitem[Lee et~al.(2019)Lee, Dosovitskiy, Bellicoso, Tsounis, Koltun, and Hutter]{AgileDynamicMotorSkills19}
Joonho Lee, Alexey Dosovitskiy, Dario Bellicoso, Vassilios Tsounis, Vladlen Koltun, and Marco Hutter.
\newblock Learning agile and dynamic motor skills for legged robots.
\newblock \emph{Sci. Robotics}, 4\penalty0 (26), 2019.

\bibitem[Li et~al.(2023)Li, Peng, Abbeel, Levine, Berseth, and Sreenath]{BipedalJumpingControl23}
Zhongyu Li, Xue~Bin Peng, Pieter Abbeel, Sergey Levine, Glen Berseth, and Koushil Sreenath.
\newblock Robust and versatile bipedal jumping control through reinforcement learning.
\newblock In Kostas~E. Bekris, Kris Hauser, Sylvia~L. Herbert, and Jingjin Yu, editors, \emph{Robotics: Science and Systems XIX, Daegu, Republic of Korea, July 10-14, 2023}, 2023.

\bibitem[Liu et~al.(2010)Liu, Yin, van~de Panne, Shao, and Xu]{SAMCONSamplingBasedContactRichMotion10}
Libin Liu, KangKang Yin, Michiel van~de Panne, Tianjia Shao, and Weiwei Xu.
\newblock Sampling-based contact-rich motion control.
\newblock In \emph{ACM SIGGRAPH 2010 Papers}, SIGGRAPH '10, New York, NY, USA, 2010. Association for Computing Machinery.
\newblock ISBN 9781450302104.

\bibitem[Long et~al.(2024)Long, Ren, Shi, Wang, Huang, Luo, and Pang]{long2024learning}
Junfeng Long, Junli Ren, Moji Shi, Zirui Wang, Tao Huang, Ping Luo, and Jiangmiao Pang.
\newblock Learning humanoid locomotion with perceptive internal model.
\newblock \emph{arXiv preprint arXiv:2411.14386}, 2024.

\bibitem[Loper et~al.(2015)Loper, Mahmood, Romero, Pons{-}Moll, and Black]{SMPL15}
Matthew Loper, Naureen Mahmood, Javier Romero, Gerard Pons{-}Moll, and Michael~J. Black.
\newblock {SMPL:} a skinned multi-person linear model.
\newblock \emph{{ACM} Trans. Graph.}, 34\penalty0 (6):\penalty0 248:1--248:16, 2015.

\bibitem[Lu et~al.(2024)Lu, Cheng, Li, Yang, Ji, Yuan, Yang, Yi, and Wang]{MobileTelevision24}
Chenhao Lu, Xuxin Cheng, Jialong Li, Shiqi Yang, Mazeyu Ji, Chengjing Yuan, Ge~Yang, Sha Yi, and Xiaolong Wang.
\newblock Mobile-television: Predictive motion priors for humanoid whole-body control.
\newblock \emph{arXiv preprint arXiv:2412.07773}, 2024.

\bibitem[Luo et~al.(2023)Luo, Cao, Winkler, Kitani, and Xu]{PHC23}
Zhengyi Luo, Jinkun Cao, Alexander Winkler, Kris Kitani, and Weipeng Xu.
\newblock Perpetual humanoid control for real-time simulated avatars.
\newblock In \emph{{IEEE/CVF} International Conference on Computer Vision, {ICCV} 2023, Paris, France, October 1-6, 2023}, pages 10861--10870. {IEEE}, 2023.

\bibitem[Luo et~al.(2024)Luo, Cao, Merel, Winkler, Huang, Kitani, and Xu]{PULSE24}
Zhengyi Luo, Jinkun Cao, Josh Merel, Alexander Winkler, Jing Huang, Kris~M. Kitani, and Weipeng Xu.
\newblock Universal humanoid motion representations for physics-based control.
\newblock In \emph{The Twelfth International Conference on Learning Representations, {ICLR} 2024, Vienna, Austria, May 7-11, 2024}. OpenReview.net, 2024.

\bibitem[Ma et~al.(2023)Ma, Farshidian, and Hutter]{ma2023learning}
Yuntao Ma, Farbod Farshidian, and Marco Hutter.
\newblock Learning arm-assisted fall damage reduction and recovery for legged mobile manipulators.
\newblock In \emph{2023 IEEE International Conference on Robotics and Automation (ICRA)}, pages 12149--12155. IEEE, 2023.

\bibitem[Makoviychuk et~al.(2021)Makoviychuk, Wawrzyniak, Guo, Lu, Storey, Macklin, Hoeller, Rudin, Allshire, Handa, and State]{IsaacGym21}
Viktor Makoviychuk, Lukasz Wawrzyniak, Yunrong Guo, Michelle Lu, Kier Storey, Miles Macklin, David Hoeller, Nikita Rudin, Arthur Allshire, Ankur Handa, and Gavriel State.
\newblock Isaac gym: High performance {GPU} based physics simulation for robot learning.
\newblock In Joaquin Vanschoren and Sai{-}Kit Yeung, editors, \emph{Proceedings of the Neural Information Processing Systems Track on Datasets and Benchmarks 1, NeurIPS Datasets and Benchmarks 2021, December 2021, virtual}, 2021.

\bibitem[Mao et~al.(2024)Mao, Zhao, Song, Shi, Ye, Zhang, Geng, Malik, Guizilini, and Wang]{UH1_24}
Jiageng Mao, Siheng Zhao, Siqi Song, Tianheng Shi, Junjie Ye, Mingtong Zhang, Haoran Geng, Jitendra Malik, Vitor Guizilini, and Yue Wang.
\newblock Learning from massive human videos for universal humanoid pose control.
\newblock \emph{arXiv preprint arXiv:2412.14172}, 2024.

\bibitem[Morimoto and Doya(1998)]{Learning2StandUp98}
Jun Morimoto and Kenji Doya.
\newblock Reinforcement learning of dynamic motor sequence: Learning to stand up.
\newblock In \emph{Proceedings. 1998 IEEE/RSJ International Conference on Intelligent Robots and Systems. Innovations in Theory, Practice and Applications (Cat. No. 98CH36190)}, volume~3, pages 1721--1726. IEEE, 1998.

\bibitem[Peng et~al.(2018)Peng, Abbeel, Levine, and van~de Panne]{DeepMimic18}
Xue~Bin Peng, Pieter Abbeel, Sergey Levine, and Michiel van~de Panne.
\newblock Deepmimic: example-guided deep reinforcement learning of physics-based character skills.
\newblock \emph{{ACM} Trans. Graph.}, 37\penalty0 (4):\penalty0 143, 2018.

\bibitem[Peng et~al.(2021)Peng, Ma, Abbeel, Levine, and Kanazawa]{AMP2021}
Xue~Bin Peng, Ze~Ma, Pieter Abbeel, Sergey Levine, and Angjoo Kanazawa.
\newblock Amp: Adversarial motion priors for stylized physics-based character control.
\newblock \emph{ACM Transactions on Graphics (ToG)}, 40\penalty0 (4):\penalty0 1--20, 2021.

\bibitem[Pinneri et~al.(2021)Pinneri, Sawant, Blaes, Achterhold, Stueckler, Rolinek, and Martius]{SampleEfficientCE21}
Cristina Pinneri, Shambhuraj Sawant, Sebastian Blaes, Jan Achterhold, Joerg Stueckler, Michal Rolinek, and Georg Martius.
\newblock Sample-efficient cross-entropy method for real-time planning.
\newblock In \emph{Conference on Robot Learning}, pages 1049--1065. PMLR, 2021.

\bibitem[Radosavovic et~al.(2024{\natexlab{a}})Radosavovic, Kamat, Darrell, and Malik]{HumanoidLocomotionChallengingTerrain24}
Ilija Radosavovic, Sarthak Kamat, Trevor Darrell, and Jitendra Malik.
\newblock Learning humanoid locomotion over challenging terrain.
\newblock \emph{arXiv preprint arXiv:2410.03654}, 2024{\natexlab{a}}.

\bibitem[Radosavovic et~al.(2024{\natexlab{b}})Radosavovic, Rajasegaran, Shi, Zhang, Kamat, Sreenath, Darrell, and Malik]{HumanoidLocomotionNextTokenPrediction24}
Ilija Radosavovic, Jathushan Rajasegaran, Baifeng Shi, Bike Zhang, Sarthak Kamat, Koushil Sreenath, Trevor Darrell, and Jitendra Malik.
\newblock Humanoid locomotion as next token prediction.
\newblock In \emph{The Thirty-eighth Annual Conference on Neural Information Processing Systems}, 2024{\natexlab{b}}.

\bibitem[Radosavovic et~al.(2024{\natexlab{c}})Radosavovic, Xiao, Zhang, Darrell, Malik, and Sreenath]{RealWorldHumanoidLocomotionScienceRobotics24}
Ilija Radosavovic, Tete Xiao, Bike Zhang, Trevor Darrell, Jitendra Malik, and Koushil Sreenath.
\newblock Real-world humanoid locomotion with reinforcement learning.
\newblock \emph{Sci. Robotics}, 9\penalty0 (89), 2024{\natexlab{c}}.

\bibitem[Sakagami et~al.(2002)Sakagami, Watanabe, Aoyama, Matsunaga, Higaki, and Fujimura]{ASIMO02}
Y.~Sakagami, R.~Watanabe, C.~Aoyama, S.~Matsunaga, N.~Higaki, and K.~Fujimura.
\newblock The intelligent asimo: system overview and integration.
\newblock In \emph{IEEE/RSJ International Conference on Intelligent Robots and Systems}, volume~3, pages 2478--2483 vol.3, 2002.

\bibitem[Schulman et~al.(2017)Schulman, Wolski, Dhariwal, Radford, and Klimov]{PPO17}
John Schulman, Filip Wolski, Prafulla Dhariwal, Alec Radford, and Oleg Klimov.
\newblock Proximal policy optimization algorithms.
\newblock \emph{arXiv preprint arXiv:1707.06347}, 2017.

\bibitem[Stelter et~al.(2021)Stelter, Bestmann, Hendrich, and Zhang]{FastAndReliable21}
Sebastian Stelter, Marc Bestmann, Norman Hendrich, and Jianwei Zhang.
\newblock Fast and reliable stand-up motions for humanoid robots using spline interpolation and parameter optimization.
\newblock In \emph{20th International Conference on Advanced Robotics, {ICAR} 2021, Ljubljana, Slovenia, December 6-10, 2021}, pages 253--260. {IEEE}, 2021.

\bibitem[Su et~al.(2024)Su, Huang, Apraez, Li, Li, Liao, Turrisi, Pontil, Semini, Wu, and Sreenath]{SymmetricLeggedLocomotion24}
Zhi Su, Xiaoyu Huang, Daniel Felipe~Ordo{\~{n}}ez Apraez, Yunfei Li, Zhongyu Li, Qiayuan Liao, Giulio Turrisi, Massimiliano Pontil, Claudio Semini, Yi~Wu, and Koushil Sreenath.
\newblock Leveraging symmetry in rl-based legged locomotion control.
\newblock In \emph{{IEEE/RSJ} International Conference on Intelligent Robots and Systems, {IROS} 2024, Abu Dhabi, United Arab Emirates, October 14-18, 2024}, pages 6899--6906. {IEEE}, 2024.

\bibitem[Tan et~al.(2016)Tan, Xie, Boots, and Liu]{HumanoidBalancing16}
Jie Tan, Zhaoming Xie, Byron Boots, and C~Karen Liu.
\newblock Simulation-based design of dynamic controllers for humanoid balancing.
\newblock In \emph{2016 IEEE/RSJ International Conference on Intelligent Robots and Systems (IROS)}, pages 2729--2736. IEEE, 2016.

\bibitem[Tao et~al.(2022)Tao, Wilson, Gou, and van~de Panne]{Learning2GetUp22}
Tianxin Tao, Matthew Wilson, Ruiyu Gou, and Michiel van~de Panne.
\newblock Learning to get up.
\newblock In Munkhtsetseg Nandigjav, Niloy~J. Mitra, and Aaron Hertzmann, editors, \emph{{SIGGRAPH} '22: Special Interest Group on Computer Graphics and Interactive Techniques Conference, Vancouver, BC, Canada, August 7 - 11, 2022}, pages 47:1--47:10. {ACM}, 2022.

\bibitem[Tassa et~al.(2012)Tassa, Erez, and Todorov]{OnlineTrajectoryOptimization12}
Yuval Tassa, Tom Erez, and Emanuel Todorov.
\newblock Synthesis and stabilization of complex behaviors through online trajectory optimization.
\newblock In \emph{2012 IEEE/RSJ International Conference on Intelligent Robots and Systems}, pages 4906--4913, 2012.

\bibitem[Tessler et~al.(2024)Tessler, Guo, Nabati, Chechik, and Peng]{MaskedMimic24}
Chen Tessler, Yunrong Guo, Ofir Nabati, Gal Chechik, and Xue~Bin Peng.
\newblock Maskedmimic: Unified physics-based character control through masked motion inpainting.
\newblock \emph{{ACM} Trans. Graph.}, 43\penalty0 (6):\penalty0 209:1--209:21, 2024.

\bibitem[Tobin et~al.(2017)Tobin, Fong, Ray, Schneider, Zaremba, and Abbeel]{DomainRandomizationSim2Real17}
Josh Tobin, Rachel Fong, Alex Ray, Jonas Schneider, Wojciech Zaremba, and Pieter Abbeel.
\newblock Domain randomization for transferring deep neural networks from simulation to the real world.
\newblock In \emph{2017 IEEE/RSJ International Conference on Intelligent Robots and Systems (IROS)}, pages 23--30, 2017.

\bibitem[Tsagarakis et~al.(2017)Tsagarakis, Caldwell, Negrello, Choi, Baccelliere, Loc, Noorden, Muratore, Margan, Cardellino, et~al.]{WalkMan17}
Nikolaos~G Tsagarakis, Darwin~G Caldwell, Francesca Negrello, Wooseok Choi, Lorenzo Baccelliere, Vo-Gia Loc, J~Noorden, Luca Muratore, Alessio Margan, Alberto Cardellino, et~al.
\newblock Walk-man: A high-performance humanoid platform for realistic environments.
\newblock \emph{Journal of Field Robotics}, 34\penalty0 (7):\penalty0 1225--1259, 2017.

\bibitem[Unitree(2024)]{UnitreeG124}
Unitree.
\newblock Unitree {G1: Humanoid Agent AI Avatar}.
\newblock 2024.
\newblock URL \url{https://www.unitree.com/g1}.

\bibitem[Vukobratovi{\'c} and Borovac(2004)]{ZeroMoment04}
Miomir Vukobratovi{\'c} and Branislav Borovac.
\newblock Zero-moment point—thirty five years of its life.
\newblock \emph{International journal of humanoid robotics}, 1\penalty0 (01):\penalty0 157--173, 2004.

\bibitem[Wang et~al.(2024)Wang, Xu, Shi, and Zhao]{guardiansasyoufall2024}
Yikai Wang, Mengdi Xu, Guanya Shi, and Ding Zhao.
\newblock Guardians as you fall: Active mode transition for safe falling.
\newblock In \emph{2024 IEEE International Automated Vehicle Validation Conference (IAVVC)}, pages 1--8. IEEE, 2024.

\bibitem[Wu et~al.(2024)Wu, Li, and Liu]{HOIHumanLevelInstruction24}
Zhen Wu, Jiaman Li, and C~Karen Liu.
\newblock Human-object interaction from human-level instructions.
\newblock \emph{arXiv preprint arXiv:2406.17840}, 2024.

\bibitem[Xue et~al.(2024)Xue, Pan, Yi, Qu, and Shi]{FullOrderSamplingBasedMPC24}
Haoru Xue, Chaoyi Pan, Zeji Yi, Guannan Qu, and Guanya Shi.
\newblock Full-order sampling-based mpc for torque-level locomotion control via diffusion-style annealing.
\newblock \emph{arXiv preprint arXiv:2409.15610}, 2024.

\bibitem[Zakka et~al.(2025)Zakka, Tabanpour, Liao, Haiderbhai, Holt, Luo, Allshire, Frey, Sreenath, Kahrs, et~al.]{MujocoPlayground25}
Kevin Zakka, Baruch Tabanpour, Qiayuan Liao, Mustafa Haiderbhai, Samuel Holt, Jing~Yuan Luo, Arthur Allshire, Erik Frey, Koushil Sreenath, Lueder~A Kahrs, et~al.
\newblock Mujoco playground.
\newblock \emph{arXiv preprint arXiv:2502.08844}, 2025.

\bibitem[Zhang et~al.(2024)Zhang, Xiao, He, and Shi]{WoCoCo24}
Chong Zhang, Wenli Xiao, Tairan He, and Guanya Shi.
\newblock Wococo: Learning whole-body humanoid control with sequential contacts.
\newblock In \emph{8th Annual Conference on Robot Learning}, 2024.

\bibitem[Zhuang and Zhao(2025)]{EmbraceCollisions25}
Ziwen Zhuang and Hang Zhao.
\newblock Embrace collisions: Humanoid shadowing for deployable contact-agnostics motions.
\newblock \emph{arXiv preprint arXiv:2502.01465}, 2025.

\bibitem[Zhuang et~al.(2024)Zhuang, Yao, and Zhao]{HumanoidParkour24}
Ziwen Zhuang, Shenzhe Yao, and Hang Zhao.
\newblock Humanoid parkour learning.
\newblock \emph{arXiv preprint arXiv:2406.10759}, 2024.

\end{thebibliography}
}

\clearpage
\begin{center}
\textbf{\textsc{\Large Appendix}}
\end{center}

\newcommand\DoToC{%
    \hypersetup{linkcolor=black}
  \startcontents
  \printcontents{}{1}{\textbf{}\vskip3pt\hrule\vskip3pt}
  \vskip7pt\hrule\vskip3pt
}

\begin{appendices}
\renewcommand{\thesubsection}{\thesection.\arabic{subsection}}

\renewcommand{\thesection}{\Alph{section}} 
\renewcommand{\sectionname}{}

\renewcommand{\thesubsection}{\Alph{section}.\arabic{subsection}}

\vspace{5pt}
\DoToC
\vspace{10pt}

\hypersetup{linkcolor=blue}
\renewcommand{\arraystretch}{1.3}
\begin{table}[ht!]
\caption{\textbf{Reward components and weights in Stage I.} Penalty rewards prevent undesired behaviors for sim-to-real transfer, regularization refines motion, and task rewards ensure successful getting up or rolling over.}
\label{tab:reward_stage1}
\centering
\resizebox{\linewidth}{!}{
\begin{tabular}{l c c}
\toprule[0.95pt]
{\scshape Term} & {\scshape Expression} & {\scshape Weight} \\
\midrule[0.6pt]
\multicolumn{3}{l}{\textit{\textbf{Penalty:}}} \\
\midrule[0.6pt]
Torque limits & $ \mathds{1}({\torque \notin [\bs{\tau}_{\min}, \bs{\tau}_{\max} ]}) $ & -0.1 \\
DoF position limits & $ \mathds{1}({\dofpos \notin [\bs{q}_{\min}, \bs{q}_{\max} ]}) $ & -5 \\
Energy & $ \lVert\ \boldsymbol{\tau} \odot \dot{\mathbf{q}} \rVert $ & -1e-4 \\
Termination & $\mathds{1}_\text{termination}$ & -500 \\
\midrule[0.6pt]
\multicolumn{3}{l}{\textit{\textbf{Regularization:}}} \\
\midrule[0.6pt]
DoF acceleration & $\lVert \dofacc \rVert_2$ & -1e-7 \\
DoF velocity & $\lVert \dofvel \rVert_2^2$ & -1e-4 \\
Action rate & $ \lVert \bs{a}_t \rVert_2^2 $ & -0.1 \\
Torque & $\lVert\torque\rVert$ & -6e-7 \\
DoF position error & $\mathds{1} (h_{\text{base}} \geq 0.8) \cdot \exp \left( -0.05 \|\dofpos - \dofpos^{\text{default}}\| \right) $ & -0.75 \\
Angular velocity & $\lVert\omega^2\rVert$ & -0.1 \\
Base velocity & $\lVert \bs{v}^2\rVert$ & -0.1 \\
Foot slip & $ \mathds{1} (\bF^{\text{feet}}_{z} > 5.0) \cdot \sqrt{\|\bs{v}_z^{\text{feet}} \|} $ & -1 \\
Feet distance reward & \makecell{
$\frac{1}{2} \Big( \exp(-100 \left| \max(\bs{d}_{\text{feet}} - \bs{d}_{\min}, -0.5) \right|)$ \\ 
$+ \exp(-100 \left| \max(\bs{d}_{\text{feet}} - \bs{d}_{\max}, 0) \right|) \Big)$
} & 2 \\
Feet orientation & $ \sqrt{\lVert \mathbf{g}_{xy}^{\text{feet}} \rVert} $ & -0.5 \\ 
Feet height reward & $\exp (-10 \cdot \bh^{\text{feet}}) $ & 2.5 \\
\midrule[0.6pt]
\multicolumn{3}{l}{\textit{\textbf{Getting-Up Task Rewards:}}} \\
\midrule[0.6pt]
Base height exp & $\exp(\bh^{base}) - 1$ & 5 \\
Head height exp & $\exp(\bh^{head}) - 1$ & 5 \\
$\Delta$ base height & $\mathds{1} (\bh_{t}^{base} > \bh^{base}_{t-1})$ & 1 \\
Feet contact forces reward & $\mathds{1} (\lVert \bF_{t}^{feet} \rVert > \lVert \bF^{\text{feet}}_{t-1}) \rVert$ & 1 \\
Standing on feet reward & $\mathds{1} \big((\lVert \bF^{\text{feet}} \rVert > 0\big)\& \big(\bh^{\text{feet}} < 0.2)\big)$ & 2.5 \\
Body upright reward & $\exp(-\mathbf{g}^{\text{base}}_z)$ & 0.25 \\
Soft body symmetry penalty & $ \left\| \mathbf{a}_{\text{left}} - \mathbf{a}_{\text{right}} \right\| $ & -1.0 \\  
Soft waist symmetry penalty & $ \left\| \mathbf{a}^{\text{waist}}\right\| $ & -1.0 \\

\midrule[0.6pt]
\multicolumn{3}{l}{\textit{\textbf{Rolling-Over Task Rewards:}}} \\
\midrule[0.6pt]
Base Gravity Error & \makecell{
$ 1 - \cos \theta_{\text{base}}$ \\ 
$\cos \theta_{\text{base}} = \frac{\mathbf{g}^{\text{base}} \cdot \mathbf{g}^{\text{target}}}{\|\mathbf{g}^{\text{base}}\| \|\mathbf{g}^{\text{target}}\|}$
} & -2 \\
Torso Gravity Error & \makecell{
$ 1 - \cos \theta_{\text{torso}}$ \\ 
$\cos \theta_{\text{torso}} = \frac{\mathbf{g}^{\text{torso}} \cdot \mathbf{g}^{\text{target}}}{\|\mathbf{g}^{\text{torso}}\| \|\mathbf{g}^{\text{target}}\|}$
} & -2 \\
Knee Gravity Error & \makecell{
$ \frac{1}{2} \Big((1 - \cos \theta_{\text{left}}^{\text{knee}}) + (1 - \cos \theta_{\text{right}}^{\text{knee}}) \big) \Big),$ \\ 
$\cos \theta^{\text{knee}} = \frac{\mathbf{g}^{\text{knee}} \cdot \mathbf{g}^{\text{target}}}{\|\mathbf{g}^{\text{knee}}\| \|\mathbf{g}^{\text{target}}\|}$
} & -2 \\
\bottomrule[0.95pt] 
\end{tabular}
}
\end{table}
\begin{figure*}[h!]
  \centering
    \includegraphics[width=\linewidth]{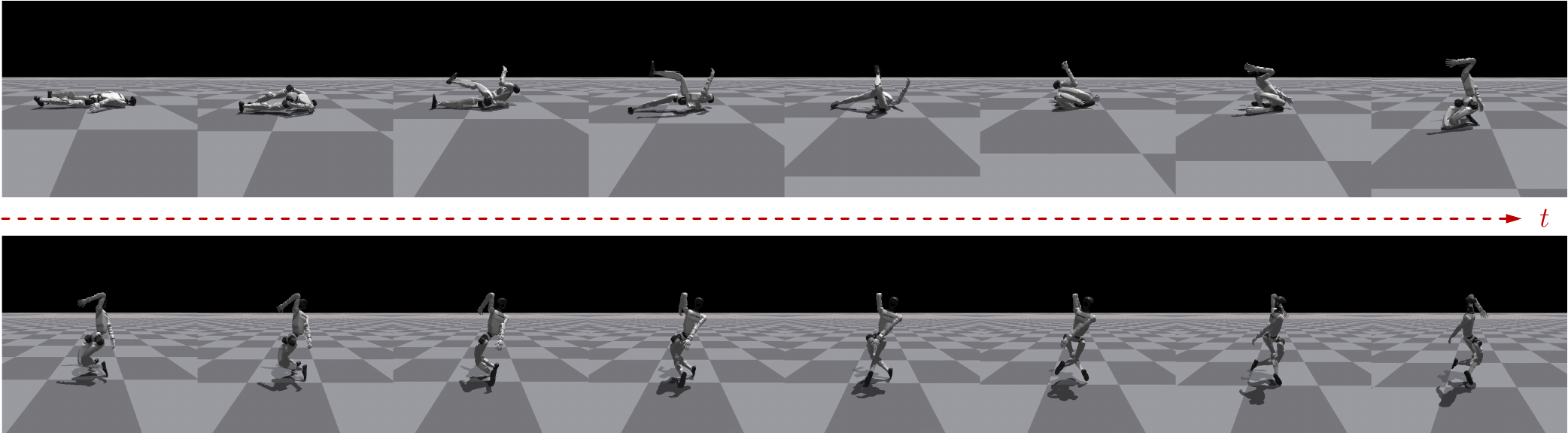}
  \caption{\textbf{Getting-up from prone pose result visualization of \citet{Learning2GetUp22}}.
  The motion generated by method~\cite{Learning2GetUp22} is highly unstable and unsafe, and it keeps jittering and jumping during the getting-up phase.
  }\label{fig:tao_case}
\end{figure*}

\section{Rewards}\label{app:rewards}

\subsection{Rewards Components in Stage I}\label{app:rewards_I}

Detailed reward components used in Stage I are summarized in \cref{tab:reward_stage1}.

\subsection{Rewards Components in Stage II}\label{app:rewards_II}
Detailed reward components used in Stage II are summarized in \cref{tab:reward_stage2}.

\begin{table}[ht!]
\caption{\textbf{Reward components and weights in Stage II.} Penalty rewards prevent undesired behaviors for sim-to-real transfer, regularization refines motion, and task rewards ensure successful whole-body tracking in real time.}
\label{tab:reward_stage2}
\centering
\renewcommand{\arraystretch}{1.3} 
\resizebox{\linewidth}{!}{
\begin{tabular}{l c c}
\toprule[0.95pt]
{\scshape Term} & {\scshape Expression} & {\scshape Weight} \\
\midrule[0.6pt]
\multicolumn{3}{l}{\textit{\textbf{Penalty:}}} \\
\midrule[0.6pt]
Torque limits & $ \mathds{1}({\torque \notin [\bs{\tau}_{\min}, \bs{\tau}_{\max} ]}) $ & -5 \\
Ankle torque limits & $ \mathds{1}({\torque^{\text{ankle}} \notin [\bs{\tau}^{\text{ankle}}_{\min}, \bs{\tau}^{\text{ankle}}_{\max} ]}) $ & -0.01 \\
Upper torque limits & $ \mathds{1}({\torque^{\text{upper}} \notin [\bs{\tau}^{\text{upper}}_{\min}, \bs{\tau}^{\text{upper}}_{\max} ]}) $ & -0.01 \\
DoF position limits & $ \mathds{1}({\dofpos \notin [\bs{q}_{\min}, \bs{q}_{\max} ]}) $ & -5 \\
Ankle DoF position limits & $ \mathds{1}({\dofpos^{\text{ankle}} \notin [\bs{q}^{\text{ankle}}_{\min}, \bs{q}^{\text{ankle}}_{\max} ]}) $ & -5 \\
Upper DoF position limits & $ \mathds{1}({\dofpos^{\text{upper}} \notin [\bs{q}^{\text{upper}}_{\min}, \bs{q}^{\text{upper}}_{\max} ]}) $ & -5 \\
Energy & $ \lVert\ \boldsymbol{\tau} \odot \dot{\mathbf{q}} \rVert $ & -1e-4 \\
Termination & $\mathds{1}_\text{termination}$ & -50 \\
\midrule[0.6pt]
\multicolumn{3}{l}{\textit{\textbf{Regularization:}}} \\
\midrule[0.6pt]
DoF acceleration & $\lVert \dofacc \rVert_2$ & -1e-7 \\
DoF velocity & $\lVert \dofvel \rVert_2^2$ & -1e-3 \\
Action rate & $ \lVert \bs{a}_t \rVert_2^2 $ & -0.1 \\
Torque & $\lVert\torque\rVert$ & -0.003 \\
Ankle torque & $\lVert\torque^{\text{ankle}}\rVert$ & -6e-7 \\
Upper torque & $\lVert\torque^{\text{upper}}\rVert$ & -6e-7 \\
Angular velocity & $\lVert\omega^2\rVert$ & -0.1 \\
Base velocity & $\lVert \bs{v}^2\rVert$ & -0.1 \\
Feet distance reward & \makecell{
$\frac{1}{2} \Big( \exp(-100 \left| \max(\bs{d}_{\text{feet}} - \bs{d}_{\min}, -0.5) \right|)$ \\ 
$+ \exp(-100 \left| \max(\bs{d}_{\text{feet}} - \bs{d}_{\max}, 0) \right|) \Big)$
} & 2 \\
Foot orientation & $ \sqrt{\lVert \mathbf{g}_{xy}^{\text{feet}} \rVert} $ & -0.5 \\
\midrule[0.6pt]
\multicolumn{3}{l}{\textit{\textbf{Tracking Rewards:}}} \\
\midrule[0.6pt]
Tracking DoF position & $ \exp \left( - \frac{(\dofpos - \dofpos^{\text{target}})^2}{4} \right) $ & 8 \\
\bottomrule[0.95pt]
\end{tabular}
}
\end{table}

\section{Additional Results}\label{app:additional_results}
\subsection{Additional Baseline Result Visualization}\label{app:baseline_result}
\cref{fig:tao_case} showcases a visualization of getting up from a prone pose generated by the baseline method~\cite{Learning2GetUp22}.
This method generates motion that is highly unstable and unsafe to deploy in the real world.
For example, its joints continuously jitter, the feet are stumbling and the body keeps jumping up.
This indicates that this baseline~\cite{Learning2GetUp22} cannot be Sim2Real.

\begin{figure*}[t!]
  \centering
    \includegraphics[width=\linewidth]{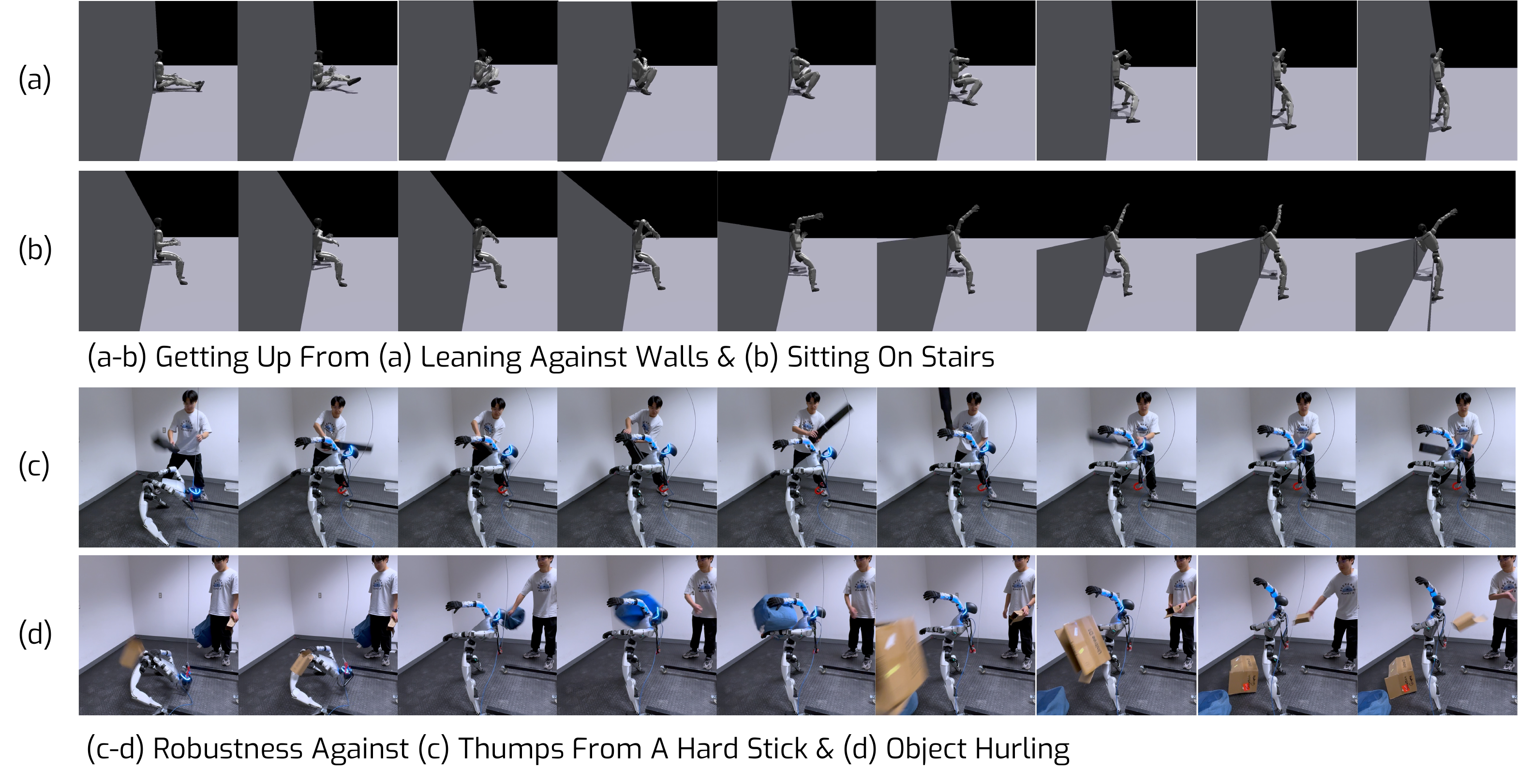}
  \caption{\textbf{Getting up from additional initial poses and against real-world external turbulence.}
  Getting up from (a) leaning against walls, (b) sitting on the stairs. Robustness to external turbulence: (c) thumps from a hard stick, and (d) object hurling.
  \ours enables the robot to get up from initial poses other than lying, and is robust against external turbulence.
  }\label{fig:robustnetss}
\end{figure*}
\subsection{Additional Results on Getting Up From Sitting On Stairs \& Leaning Against Walls}\label{app:sitting_results}
\cref{fig:robustnetss}(a) and (b) show the simulation results of getting up from additional initial postures:
sitting on the stairs and leaning against the wall.
The results show that \ours can be generalized to more diverse initial postures in addition to lying on the ground.
Besides, we find that the convergence is $\sim4\times$ faster than getting up from lying.
We argue that getting up from sitting or leaning is easier because of the additional support from the ground, chairs, or walls.

\subsection{Additional Robustness Test Against External Turbulence}\label{app:robust_results}
\cref{fig:robustnetss}(c) and (d) show the robustness test against external turbulence:
thumps from a hard stick and object hurling.
The results show that \ours getting-up policy is practically robust against certain external turbulence in the real world.

\section{Training Details}\label{app:training_details}
In Stage I, we train the discovery policy $f$ for overall 5B simulation steps, and 20K simulation steps for the Stage II deployable policy $\pi$.
Each stage uses a regularization curriculum within, an implementation detail common to policy learning in legged locomotion literature. 
All training is conducted on Isaac Gym~\cite{IsaacGym21}, and we train our policies using 4,096 paralleled environments on a single NVIDIA RTX 4090 or L40S GPU.
For the getting-up task, we slow down the discovered trajectory to 8 seconds ($8\times$).
We also tried $4\times$ and $10\times$.
$4\times$ leads to large torques and DoF velocities, and $10\times$ does not converge. 
For the rolling-over task, the trajectory is slowed down to 4 seconds
(selected through trials similar to the getting-up task).
We use flat terrains in Stage I and varied terrains during Stage II, involving flat, rough, and slope terrains.
We follow previous works~\cite{OmniH2O24,ExtremeParkour24,HumanoidParkour24} to apply varied dynamics randomization, such as base center of mass (CoM) offset and control delay.

\end{appendices}

\end{document}